\documentclass[11pt]{article}

\usepackage[preprint]{acl}

\usepackage{times}
\usepackage{latexsym}
\usepackage{graphicx}
\usepackage{amssymb} 
\usepackage{amsmath}    
\usepackage{booktabs}   
\usepackage{multirow}   
\usepackage{subcaption}
\usepackage[T1]{fontenc}

\usepackage[utf8]{inputenc}

\usepackage{microtype}

\usepackage{inconsolata}

\usepackage{hyperref}
\usepackage{url}
\usepackage{float}
\usepackage{xcolor}
\usepackage{tcolorbox}

\tcbuselibrary{breakable}  
\tcbuselibrary{skins}      

\definecolor{color1}{HTML}{E74C3C} 
\definecolor{color2}{HTML}{2ECC71}
\definecolor{color3}{HTML}{3498DB} 
\definecolor{color4}{HTML}{F39C12} 
\definecolor{color5}{HTML}{34495E}

\definecolor{color6}{HTML}{2ECC71} 
\definecolor{color7}{HTML}{E74C3C} 
\definecolor{color8}{HTML}{8B4513}

\definecolor{color9}{HTML}{E74C3C} 
\definecolor{color10}{HTML}{2ECC71} 
\definecolor{color11}{HTML}{8E44AD}

\definecolor{color12}{HTML}{1f77b4} 
\definecolor{color13}{HTML}{ff7f0e} 
\definecolor{color14}{HTML}{2ca02c} 
\definecolor{color15}{HTML}{d62728} 
\title{More Bang for the Buck: Process Reward Modeling \\ with Entropy-Driven Uncertainty}

\author{
  \textbf{Lang Cao \quad Renhong Chen \quad Yingtian Zou \quad Chao Peng} \\
  \textbf{Huacong Xu \quad Yuxian Wang \quad Wu Ning \quad Qian Chen \quad Mofan Peng} \\ 
  \textbf{Zijie Chen \quad Peishuo Su \quad Yitong Li} \\
  Huawei Technologies Co., Ltd., China \\
  \small{\{caolang1019, f.w.lrank\}@gmail.com}
}

\begin{document}
\maketitle
\begin{abstract}
We introduce a novel entropy-driven training framework, Entropy-Driven Uncertainty Process Reward Model (EDU-PRM), for modeling process reward that enables dynamic and uncertainty-aligned segmentation of complex reasoning steps.
Unlike previous Process Reward Models (PRMs) that rely on static partitioning or human labeling, EDU‑PRM automatically anchors step boundaries at tokens with high predictive entropy, which can effectively capture intrinsic logical transitions and facilitating efficient exploration of diverse reasoning paths.
On the ProcessBench benchmark, EDU-PRM outperforms strong public PRM baselines, such as Math-Shepherd PRM and Omega PRM, and EDU-PRM achieves comparable results to the SOTA Qwen2.5-Math-PRM while using only $1.5\%$ of its publicly reported process-level training data.
Furthermore, by leveraging our proposed EDU sampling strategy, we observe accuracy boosts from $64.7\%$ to $67.3\%$ for reasoning tasks, accompanied by a reduction of $32\%$ in token usage.
These findings underscore the potential of EDU-PRM as a scalable and annotation-efficient paradigm for process supervision in mathematical reasoning, paving the way for more efficient and robust approaches to complex mathematical problem solving.
\end{abstract}

\section{Introduction}

Large Language Models (LLMs), such as GPT-4o~\citep{openai2024gpt4ocard} and Deepseek-V3~\citep{deepseekai2025deepseekv3technicalreport}, have achieved remarkable performance in a wide range of tasks.
Despite these successes, LLMs still struggle with complex multi-step reasoning problems, where verifying each intermediate reasoning step is essential to producing reliable solutions~\citep{wei2023chainofthoughtpromptingelicitsreasoning}.
To address these challenges, recent approaches adopted reinforcement learning~\citep{murphy2025reinforcementlearningoverview} with reward models, moving from supervision focused solely on final answers to more granular and step-level evaluations using LLM judges.

Process Reward Models (PRMs)~\citep{lightman2023letsverifystepstep} present a significant step forward by providing stepwise feedback, improving both the reliability and the interpretability of the model reasoning.
However, the practice of PRMs introduces two critical challenges.
First, defining what constitutes a ``correct'' intermediate step is often ambiguous, and obtaining step-level data is difficult and requiring large-scale human annotation, e.g. the PRM800K dataset~\citep{lightman2023letsverifystepstep}, is time-consuming and costly.
Recent methods, such as Qwen2.5-PRM~\citep{zheng2025processbenchidentifyingprocesserrors,zheng2023judgingllmasajudgemtbenchchatbot}, employ LLM-based judgment or Monte Carlo estimation~\citep{xie2024montecarlotreesearch,zhang2024restmctsllmselftrainingprocess} to scale supervision, however, these approaches still require substantial computational resources.
Second, the reliability of intermediate evaluation remains limited.
PRMs can be ``cheating'', as high step scores do not guarantee a correct final answer~\citep{deepseekai2025deepseekv3technicalreport}, which undermines the effectiveness of stepwise supervision and poses a significant barrier to robust reasoning.

To overcome these challenges, we propose \textbf{Entropy-Driven Uncertainty Process Reward Model (EDU-PRM)}, a novel framework for scalable and efficient step-level supervision without the need for expensive human or LLM annotations.
Our approach leverages entropy-driven sampling to automatically generate diverse and informative intermediate reasoning steps.
Furthermore, by explicitly modeling uncertainty, EDU-PRM improves the alignment between stepwise evaluation and final answer correctness, thereby mitigating the ``cheating'' issue.

We summarise our contributions as follows.

\textbf{EDU Sampling for PRM Training.}
We propose an entropy-driven uncertainty (EDU) sampling strategy to automatically generate diverse and informative step-level data.
Unlike prior PRMs such as Qwen2.5-Math-PRM, which require LLM or symbolic supervision at \emph{every} intermediate step, EDU-PRM only relies on final-answer correctness.
Fragment-level rewards are inferred automatically via Monte Carlo aggregation, without any step-wise human or LLM labeling.

\textbf{Reliable Stepwise Supervision.}
By assigning soft Monte Carlo rewards to entropy-aligned fragments, PRMs trained with EDU sampling achieve substantially better alignment between stepwise evaluation and final answer correctness, reducing the ``cheating'' phenomenon, where high process scores fail to yield correct final answers.

\textbf{Efficient and Accurate Solution Generation.}
Applying EDU sampling during inference leads to comparable or higher accuracy than conventional high-temperature sampling with substantially lower token budgets, up to $32\%$ fewer tokens on MATH and OLY benchmarks.

In summary, EDU-PRM enables scalable, annotation-efficient, and reliable step-level supervision for complex reasoning tasks.

\begin{figure*}[t!]
    \centering
    \includegraphics[width=\textwidth]{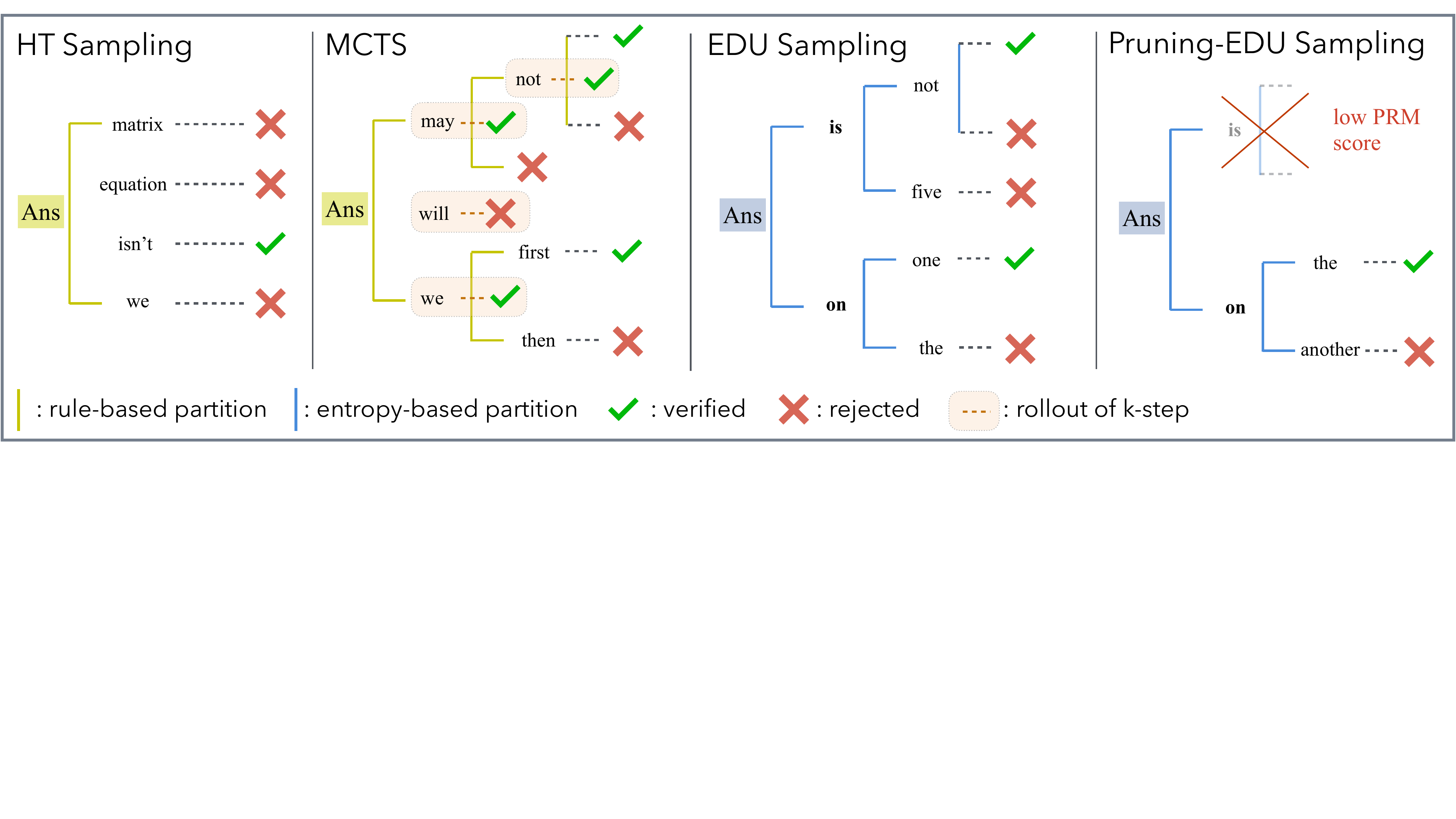}
    \caption{Comparison of sampling methods in Process Reward Models (PRMs). 
    (Left first) \textbf{High Temperature (HT) sampling} selects the best answer from \(N\) candidates but incurs \(\mathcal{O}(N)\) cost. 
    (Left second) \textbf{OmegaPRM} uses MCTS for tree search but relies on rule-based partitioning. 
    (Right first and second) \textbf{EDU Sampling (Ours)} dynamically branches at high-entropy tokens (red segments, e.g., ``is'', ``on''), which act as logical anchors. 
    The \textbf{Pruning-EDU} strategy further optimizes this by pruning branches with low PRM scores early (indicated by the ``rejected'' path), effectively concentrating computation on promising reasoning trajectories.}
    \label{fig:method_compare}
\end{figure*}

\section{Related Works}\label{sec:relatedworks}
Methods for evaluating LLM outputs have evolved from early rule-based heuristics to sophisticated model-based reward frameworks.
Initial approaches~\citep{mu2024rulebasedrewardslanguage} relied on keyword matching, which limited their generalizability when domain transferring.
The LLM-as-judge paradigm~\citep{zheng2023judgingllmasajudgemtbenchchatbot} enabled self-evaluation but introduced self-verification biases, as well as increased computational costs~\citep{wang2023selfconsistencyimproveschainthought}.
Output-Reward Models (ORMs;~\citealp{wang2024interpretablepreferencesmultiobjectivereward,yuan2024freeprocessrewardsprocess,luo2024improve}) assign scores to final outputs based on human annotation.
However, ORMs neglect intermediate reasoning steps, risking misjudgment when flawed processes yield correct results.

To address this, Process Reward Models~\citep{lightman2023letsverifystepstep,zhang2025lessonsdevelopingprocessreward} score reasoning chains at step level, using either soft labels (LLM-generated scores) or hard labels (expert binary judgments).
Soft labels enable scalable annotation but may introduce bias, while hard labels offer reliability at a higher cost.
PRMs improve reliability in tasks such as mathematical reasoning by penalizing erroneous intermediate steps.

However, key challenges remain, such as the difficulty of obtaining high-quality labels and the limited effectiveness of current PRM approaches~\citep{deepseekai2025deepseekr1incentivizingreasoningcapability,wu2024metarewardinglanguagemodelsselfimproving,sun2024easytohardgeneralizationscalablealignment,yin2025dynamicgeneralizableprocessreward}.
Math-Shepherd PRM~\citep{wang2024mathshepherdverifyreinforcellms} employs a two-stage process.
The base model generates solution traces via self-consistency sampling, and a symbolic checker verifies answers and propagates binary labels to intermediate steps.
This automatic chain annotation reduces manual effort and supports efficient PRM training.
Omega PRM~\citep{luo2024improvemathematicalreasoninglanguage} frames the problem-solving procedure as a search tree problem, using Monte-Carlo Tree Search to decompose tasks and explore promising branches.
The PRM predictions guide tree exploration and serve as rewards during policy optimization, enhancing exploration efficiency and reasoning capability.

\paragraph{Uncertainty and Entropy in Reasoning}
Recent studies have leveraged entropy and uncertainty primarily for regularization, verification, and data construction.
Entropy-regularized approaches~\citep{zhang2025entropyregularizedprocessrewardmodel} apply global penalties to encourage diversity but act as passive statistical constraints without guiding step-wise segmentation.
Similarly, uncertainty metrics have been employed for step-wise verification~\citep{ye2025uncertaintyawarestepwiseverificationgenerative} and automated data construction~\citep{han2025uncertaintybasedmethodsautomatedprocess}.
While effective for filtering unreliable steps or aggregating final answers, these methods operate primarily in a \textit{post-hoc} manner—monitoring or selecting outputs after generation rather than actively steering the reasoning trajectory.

In contrast, our approach utilizes entropy as an active \textit{control signal} to dynamically segment reasoning steps and trigger branching.
Instead of relying on static regularization or post-hoc filtering, we use adaptive entropy thresholds to structure reasoning process in real-time.
This enables fine-grained, context-sensitive exploration that integrates seamlessly with PRM and Best-of-N strategies, providing robust guidance against local optima.

\section{Methodology}

As discussed in Section~\ref{sec:relatedworks}, existing PRMs still face several critical challenges, such as the difficulty of obtaining high-quality labels and the limited effectiveness of predicting final answers.
In particular, many conventional PRMs rely on superficial textual cues such as blank lines or punctuation to segment reasoning steps and to assign rewards.
However, these heuristics fail to capture the underlying logical transitions in complex solution traces, resulting in suboptimal supervision and limited generalization.

Recent advances in reasoning with LLMs have highlighted the importance of stepwise exploration during solution generation.
In particular, Chain-of-Thought (CoT) Decoding~\citep{wang2024chainofthoughtreasoningprompting} demonstrates that branching at token positions where the model exhibits uncertainty, specifically the probability gap between the top-$1$ and top-$2$ candidates is small, can reveal alternative reasoning paths and improve overall solution quality.
Building on this insight, studies further establish that high-entropy tokens serve as natural anchors for meaningful exploration~\cite{cheng2025reasoningexplorationentropyperspective}.
These tokens often correspond to logical pivots or transitions in the reasoning process, making them ideal candidates for step segmentation and branching.

Motivated by these findings, we propose Entropy-Driven Uncertainty Process Reward Model (EDU-PRM).
By dynamically identifying and branching at positions of maximal uncertainty, our EDU-PRM is able to generate logically coherent, diverse, and informative step-level data.
This approach not only enhances the quality of process supervision but also reduces reliance on manual annotation and rigid heuristics, paving the way for more robust and scalable reward modeling.


\subsection{Entropy-Driven Uncertainty Sampling}\label{sec:edu_sampling}
Token entropy often used to quantify the uncertainty in predicting the next token at each decoding step.
High entropy indicates that the probability distribution over possible next tokens is more dispersed, reflecting greater ambiguity or indecision.
In contrast, low entropy indicates the model is confident, with most probability mass assigned to a single token.

EDU sampling leverages these high-entropy tokens as \emph{uncertainty anchors}, guiding the segmentation of reasoning steps to better reflect the underlying logical structure of the solution trace, rather than relying on superficial textual cues.

Formally, we apply softmax function to the output logits of an autoregressive model at each decoding step $t$, yielding a probability distribution \(P_{v}\) over possible next tokens $v$~\citep{kwon2023efficientmemorymanagementlarge,aminabadi2022deepspeedinferenceenablingefficient}.
Then, the entropy at $t$ is calculated as:
\begin{equation}
    H_{t} = -\sum_{v} P_{v} \cdot \log\left( P_{v} + \epsilon \right)
\end{equation}
where \(\epsilon\) is a small constant for numerical stability.

As illustrated in Figure~\ref{fig:method_compare}, our EDU sampling workflow consists of two main stages: 1) entropy-based anchor detection and branching, and 2) fragment-level evaluation and labeling.

\paragraph{EDU Sampling at Uncertainty Anchors}
We define position $t$ as an \textbf{uncertainty anchor} when \(H_{t}\) exceeds a threshold \(\tau\). 

To balance solution diversity and quality, at each uncertainty anchor, EDU sampling branches into $2$ using top-$2$ logits,\footnote{Experiments with top-$3$ and other schemes yielded similar results.} and it then generate subsequent tokens greedily (i.e. \(\arg\max_v \mathbf{P}_v\)) until the next uncertainty anchor is reached.
This strategy efficiently samples alternative reasoning paths without excessive computational overhead.
To avoid artifacts caused by specific structural tokens (e.g., opening parentheses or brackets), we exclude tokens in the symbol set \(\mathcal{S}\) (see Appendix~\ref{app:mathcal_s}) from entropy calculations.

\paragraph{Monte Carlo Estimation Scoring}
After performing the EDU sampling, model generates a binary tree, where each branch is segmented into fragments by uncertainty anchors.
To score each fragment, we assign a correctness label $(0, 1)$ based on the final solution’s validity using Monte Carlo Estimation (MCE;~\citealp{katzgraber2011introductionmontecarlomethods}).
This fragment-level scoring approach enables a fine-grained assessment of reasoning steps, as shown in Figure~\ref{fig:method_compare}, where each segment is mapped to its corresponding correctness label.

\subsection{Entropy-Driven Uncertainty PRM}

We perform the proposed EDU sampling workflow to construct the corpus for the EDU-PRM training, where each instance consists of a triple, i.e. a question, a solution (or a solution fragment), and an associated label indicating the correctness of the solution.
We then train EDU-PRM via a classification-oriented cross-entropy loss, $\mathcal{L} = -\frac{1}{N}\sum_{i=1}^{N} y_{i}\log p_{i}$, where \(N\) is the number of examples, \(y_{i}\) are the target label, and \(p_{i}\) denotes the predicted probabilities from logits.
Note that our methods do not introduce human efforts to segment or to label intermediate reasoning steps, and we show the effectiveness of the uncertainty anchor-based segmentation methods in the following experimental sections.

\section{Experiments}
We demonstrate our proposed EDU-PRM using two setups, a direct evaluation over PRM benchmarks and evaluation by applying PRMs as a BoN results selector over a series of math reasoning tasks.
In addition, we also experiment with the proposed EDU sampling strategy focusing not only on accuracy but also on token efficiency, comparing with the traditional high-temperature (HT) sampling method.

\subsection{Implementations of EDU-PRM}\label{sec:exp_impl}

We follows the implementation of Math-Shepherd PRM~\citep{wang2024mathshepherdverifyreinforcellms} and Omega PRM~\citep{luo2024improvemathematicalreasoninglanguage} with consistent experimental settings and parameter configurations to train all models.

For detailed model training, we use data from the MATH training set~\citep{hendrycks2021measuringmathematicalproblemsolving}, selecting $7,500$ problems as the base query set and sampling up to $100$ candidate solutions per problemusing the EDU sampling (token-level predictive entropy threshold = $1.0$)
These form a training set of approximately 1.42M instances, with a label distribution of $52\%$ hard and $48\%$ soft labels.
We use the entropy threshold of $1.0$ as it empirically yields an optimal balance between segmentation granularity and search efficiency.

We experiment with two model size, Qwen2.5-72B-Base and Qwen2.5-7B-Base~\citep{qwen2025qwen25technicalreport}.
All the details of the training frameworks, dataset statistics, and inference hyperparameters are listed in Appendix~\ref{app:dataset}, and the prompts used for solution verification are also provided in Appendix~\ref{app:evaluation_prompt}.

\subsection{Evaluation Benchmarks and Comparison Baselines}
\label{subsec:datasets_baselines}

We evaluate the effectiveness of PRMs from two aspects.
On one hand, we directly evaluate the accuracy of PRMs using a well-established RPM benchmark, processBench~\citep{zheng2025processbenchidentifyingprocesserrors}, where PRMs aim to predict whether the response is correct or not.
On the other hand, we perform a Best-of-$N$ (BoN) evaluation of RPMs on real-world math reasoning tasks.
In this setting, PRMs aim to select the correct answers from $N$ response candidates.
We select a range of math benchmarks with different difficulties, including OlympiaBench (OLY)~\citep{he2024olympiadbenchchallengingbenchmarkpromoting}, MATH~\citep{hendrycks2021measuringmathematicalproblemsolving}, GSM8K~\citep{cobbe2021trainingverifierssolvemath}, and CollegeMath~\citep{tang2024mathscalescalinginstructiontuning}, and for each query, we generate $128$ candidate solutions using Qwen2-7B-Instruct~\citep{yang2024qwen2technicalreport}.

We compare with sota PRMs, including Math-Shepherd-Mistral-7B-PRM~\citep{wang2023math}, Qwen2.5-Math-7B-PRM800K, Qwen2.5-Math-PRM-7B, Qwen2.5-Math-PRM-72B, and Qwen2.5-Math-RM-72B~\citep{yang2024qwen25mathtechnicalreportmathematical}.
Note that the open-sourced versions of these baselines are trained on much larger datasets than ours.
For fair comparison, we re-implement these baselines based on the same data and base models as EDU-PRM, except the Qwen2.5-Math-PRM series.
We report the performance of the original version of Qwen2.5-Math-PRMs as strong sota baselines.

\begin{figure}[t!]
    \centering
    \includegraphics[width=0.85\linewidth]{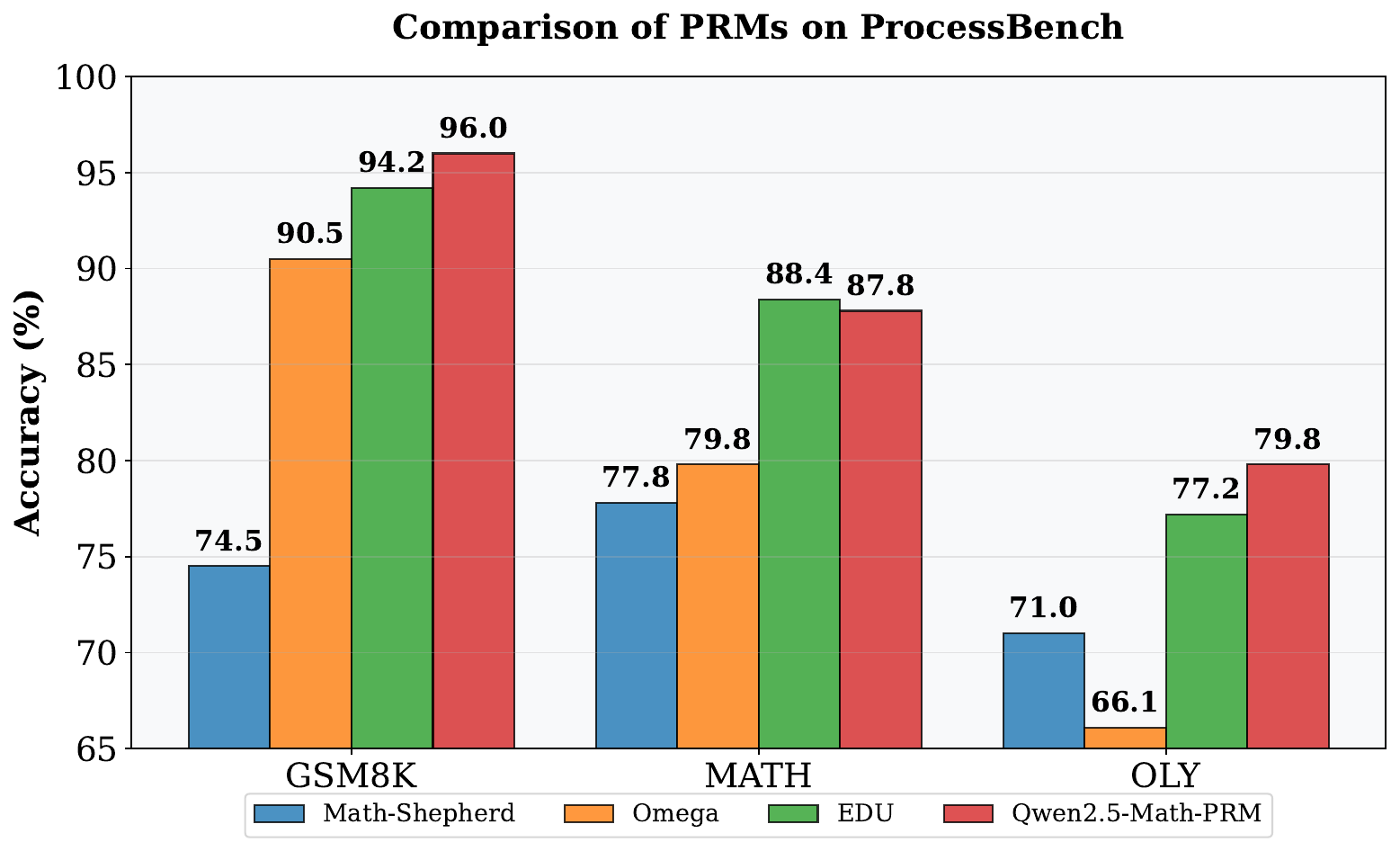}
    \caption{Accuracy comparison on ProcessBench for four 72B-parameter PRMs: \textcolor{color12}{Math-Shepherd PRM}, \textcolor{color13}{Omega PRM}, \textcolor{color14}{EDU PRM}, and \textcolor{color15}{Qwen2.5-Math-PRM}. As a competitive PRM method, our proposed \textcolor{color14}{EDU PRM} attains the highest accuracy on the MATH test dataset. On GSM8K and OLY datasets, \textcolor{color14}{EDU PRM} matches the performances of \textcolor{color15}{Qwen2.5-Math-PRM}.} 
    \label{fig:comparison_method_process_bench}
\end{figure}

\subsection{Accuracy Evaluation of PRM}

Figure~\ref{fig:comparison_method_process_bench} demonstrates that EDU-PRM-72B achieves outstanding performance in solution correctness judgment across multiple benchmarks. On the MATH dataset, EDU-PRM-72B attains the highest judgment accuracy of $88.4\%$, outperforming Qwen-2.5-math-PRM-72B ($87.8\%$) by a margin of $0.6\%$. Additionally, EDU-PRM-72B exhibits robust judgment accuracy on GSM8K ($94.2\%$) and OlympicBench ($77.2\%$), further highlighting its effectiveness in verifying mathematical solutions. Notably, EDU-PRM-72B consistently surpasses Math-Shepherd PRM and Omega PRM across all evaluated benchmarks. Detailed experimental results are provided in Appendix~\ref{app:process_bench}.

It is worth noting that, as shown in Table~\ref{tab:model_performance}, the 7B models generally exhibit lower recall and F1 scores compared to their 72B counterparts. 
This performance gap is primarily attributed to the limited capacity of smaller models in handling the imbalanced label distribution inherent in the training data, as well as their weaker reasoning capabilities on complex tasks. 
However, our 72B models consistently achieve state-of-the-art results under the same data conditions, demonstrating the scalability of our approach.

\begin{figure*}[t!]
    \centering
    \includegraphics[width=\linewidth]{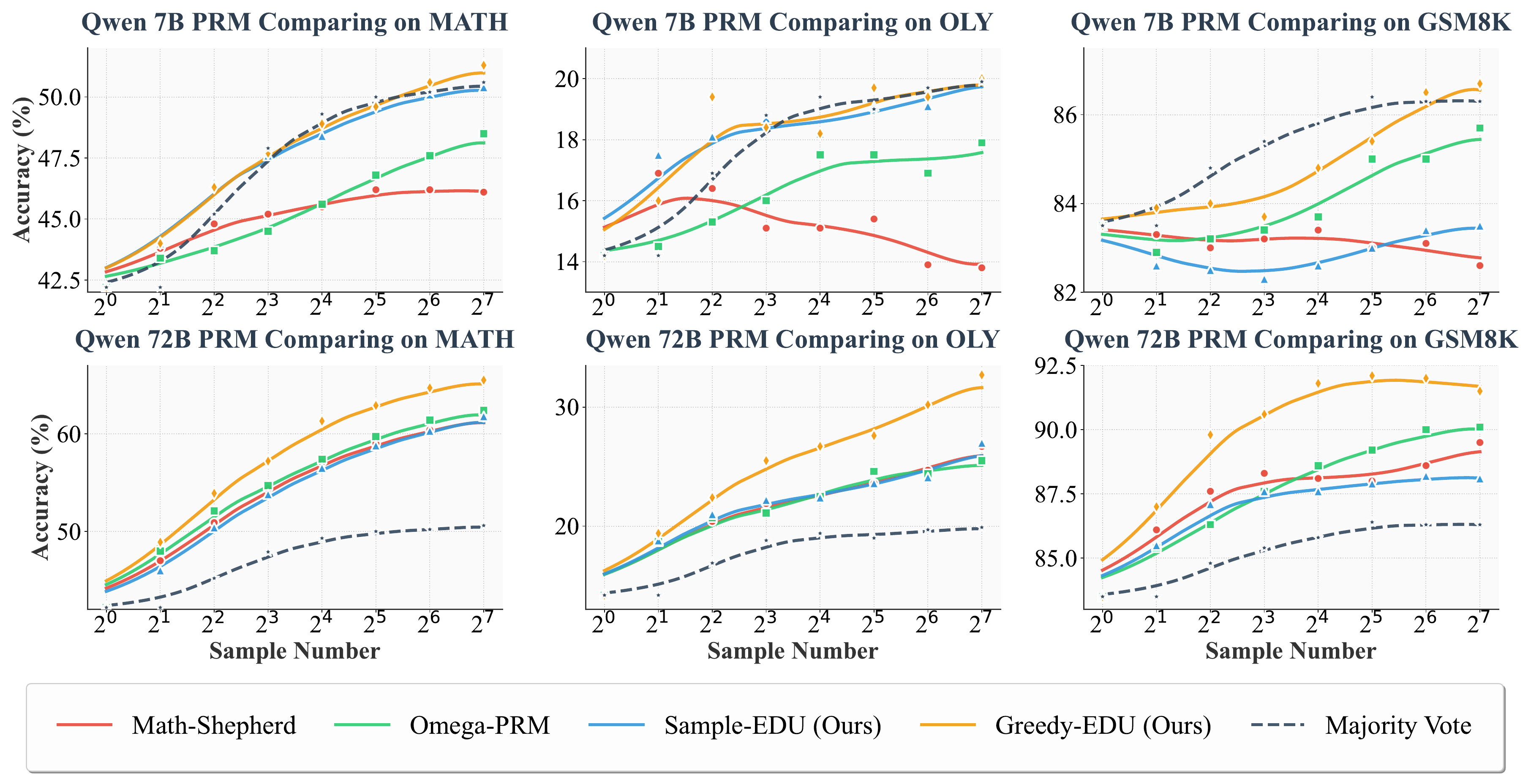}
    \caption{Comparison of PRM performance on the MATH, OLY, and GSM8K benchmarks for Qwen 7B and 72B models. Evaluated methods: \textcolor{color1}{Math-Shepherd}, \textcolor{color2}{Omega-PRM}, \textcolor{color3}{Sample-EDU}, \textcolor{color4}{Greedy-EDU}, \textcolor{color5}{Majority Vote} serves as a non‑PRM baseline. Markers show raw scores; curves are Gaussian-smoothed (trend visualisation only). \textcolor{color4}{\textbf{Greedy-EDU}} consistently leads or matches the best results across datasets and model scales.}
    \label{fig:comparison_method_of_prm}
\end{figure*}

\subsection{Evaluating PRMs via BoN}\label{sec:method-section}

Figure~\ref{fig:comparison_method_of_prm} summarises the performance of different models across three datasets, highlighting the superior results of Greedy-EDU PRM (i.e. EDU-7B and EDU-72B respectivly).
We observed that EDU-72B achieves up to a $3.7\%$ lead on MATH and a $5.7\%$ lead on OLY consistently across different sampling sizes, compared with SOTA baselines.
When compared with majority voting, usually considered as a strong baseline of BoN, our PRM-based method can consistently achieve better accuracy of response selection, especially when the model size increases.
Full experimental results are detailed in Table~\ref{tab:compare_PRM_result}.

\begin{figure*}[t!]
    \centering
    \includegraphics[width=0.9\linewidth]{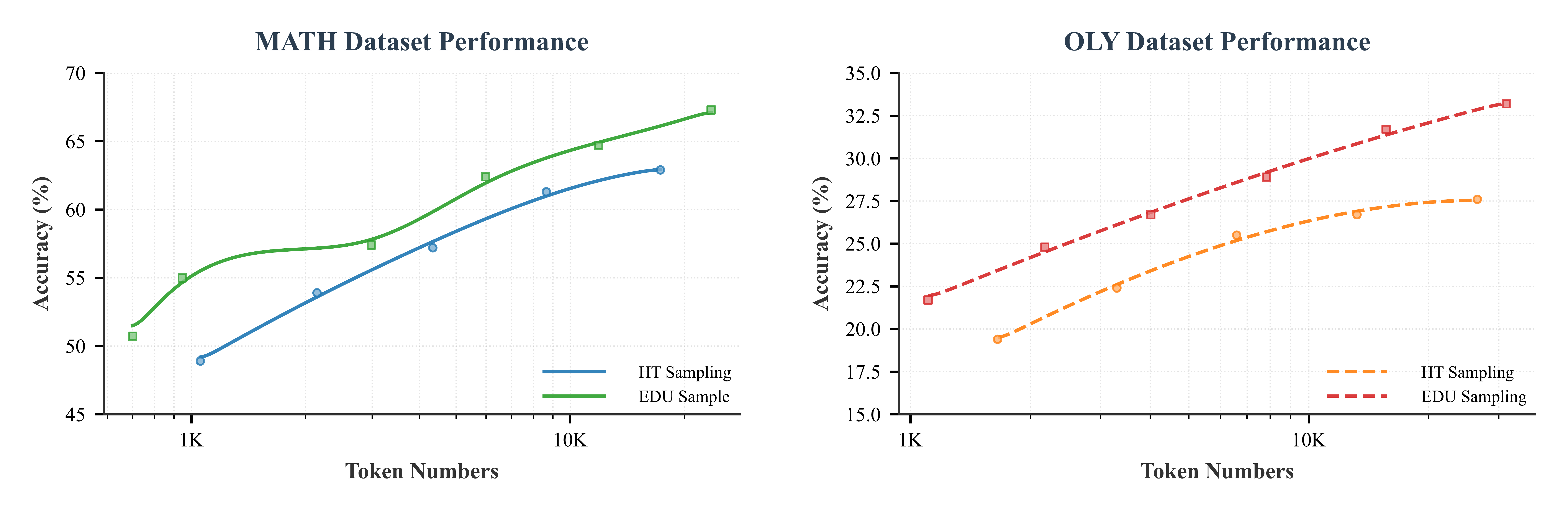}
    \caption{Comparison of sample strategies under the EDU‑PRM 72B model on the MATH and OLY test sets: High‑Temperature (HT) Sampling, EDU Sampling. Markers denote raw measurements; curves are Gaussian‑smoothed trends. Points nearer the upper‑left frontier indicate a better accuracy–token trade‑off. It can be observed that on both the OLY and MATH test sets, EDU Sampling achieves an overall higher accuracy compared to HT Sampling while consuming fewer tokens.}
    \label{fig:comparison_acc_per_token}
\end{figure*}

\subsection{Sampling Strategy Comparison: EDU Sampling vs. HT Sampling}
After demonstrating the superior performance of EDU-PRM, we further investigate the accuracy and token efficiency of our EDU sampling strategies during candidate inference.
Specifically, we adopt the BoN evaluation setup while using EDU sampling instead of the traditional HT Sampling (temperature = $0.7$).

Experimental results on the MATH and OLY test sets (see Figure~\ref{fig:comparison_acc_per_token}) show that EDU sampling consistently outperforms HT sampling in both accuracy and token efficiency.
On MATH, EDU sampling achieves $57.4\%$ accuracy with $2,988$ tokens, while HT sampling achieves $57.2\%$ accuracy with $4,338$ tokens on average.
On OLY, EDU sampling attains $21.7\%$ accuracy with $1,107$ tokens, compared to $19.4\%$ of HT sampling with $1,655$ tokens.

Both methods initially show increasing accuracy with more tokens, however at higher token counts, EDU sampling maintains a steep upward trajectory in accuracy, while HT sampling improves plateaus, indicating diminishing returns.
This highlights EDU sampling’s superior capability to leverage additional tokens for sustained accuracy gains.

Overall, these results indicate that the EDU sampling not only achieves higher accuracy but also utilizes tokens more efficiently, making it a preferable strategy for mathematical reasoning tasks under computational constraints.

\begin{figure*}[t!]
    \centering
    \includegraphics[width=0.9\linewidth]{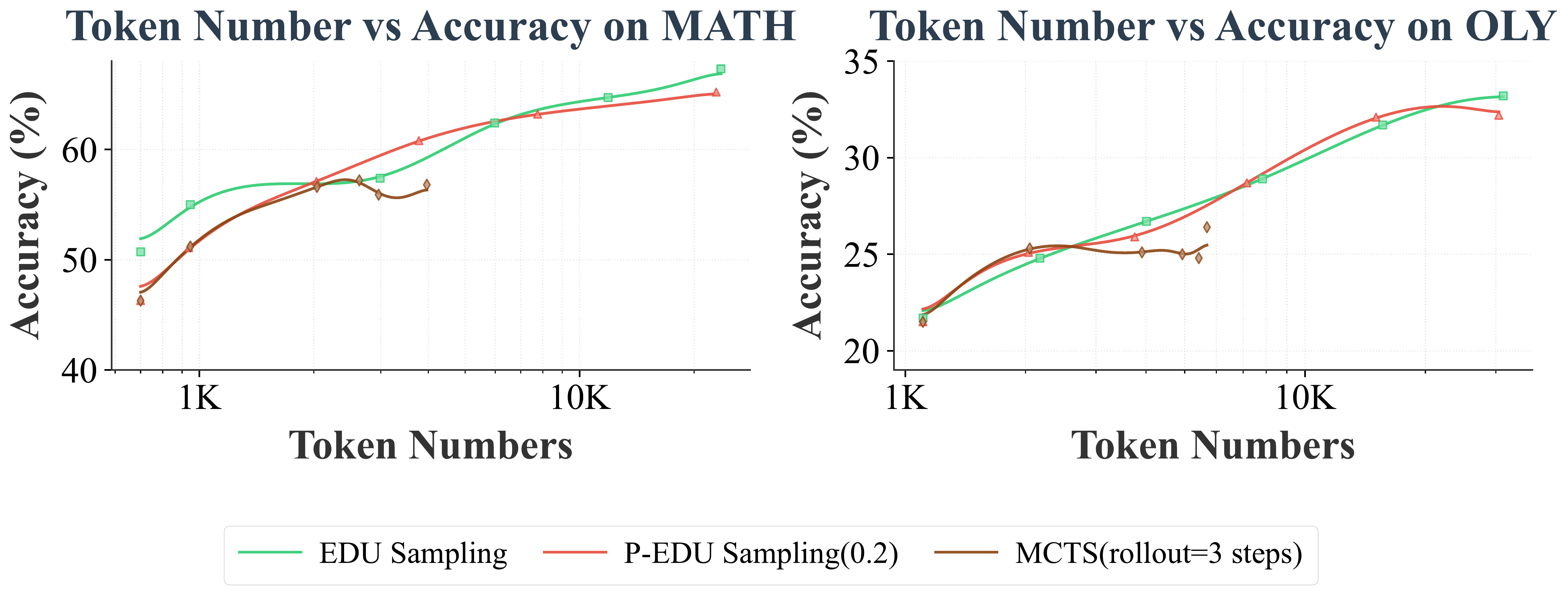}
    \caption{Comparison of sample strategies under the EDU‑PRM 72B model on the MATH and OLY test sets: \textcolor{color6}{EDU Sampling}, \textcolor{color7}{P-EDU Sampling} (with a threshold of $0.2$), and \textcolor{color8}{MCTS} (with exploration depth not exceeding $3$ steps rollout). Markers denote raw measurements; curves are Gaussian‑smoothed trends. The x-axis represents token counts, and the y-axis represents accuracy (\%). Points nearer the upper-left frontier indicate a better accuracy–token trade-off. \textcolor{color7}{P-EDU Sampling} achieves a measurable lead on both the OLY and MATH test sets, yet \textcolor{color6}{EDU Sampling} exhibits a more pronounced advantage under high token counts across both test sets. Detailed raw numerical results for these comparisons are provided in Table~\ref{tab:math_oly_combined}.
}
    \label{fig:comparison_acc_per_token_mcts_pedu}
\end{figure*}

\subsection{Efficiency and Scalability: Comparing EDU Variants with MCTS}

To further explore the trade-off between solution quality and generation efficiency, we propose \textbf{Pruning-EDU (P-EDU)} as a token-efficient variant of our framework and compare it against the established \textbf{Monte Carlo Tree Search (MCTS)} baseline.
Specifically, P-EDU sampling applies a pruning threshold of $0.2$ to filter out low-confidence branches. 
We report this threshold because it achieves the best trade-off between token efficiency and accuracy; setting it too high risks pruning correct answers early, whereas a lower threshold fails to significantly reduce token usage.
In contrast, MCTS leverages forward-looking exploration with a rollout depth of $3$ steps.
By simulating future reasoning steps, it can make more informed decisions about which current paths are worth pursuing, rather than relying solely on immediate scores.

Table~\ref{tab:math_oly_combined} and Figure~\ref{fig:comparison_acc_per_token_mcts_pedu} summarize the distinct performance profiles of these strategies on both the MATH and OLY test sets.
The results highlight the superior scalability of our EDU-based methods.
EDU sampling’s accuracy steadily increases with more tokens, dominating the high-accuracy frontier. 
Simultaneously, P-EDU sampling achieves a balanced trade-off, reaching $32.1\%$ accuracy at $15,050$ tokens on OLY—comparable to EDU sampling in the mid-token range—benefited from the effective pruning of low-confidence paths.
On the MATH dataset, while MCTS performs competitively in the low-token regime (achieving $51.2\%$ accuracy at $946$ tokens), it hits a distinct performance ceiling.
Unlike EDU methods, further increasing the token budget for MCTS does not yield proportional accuracy gains, as its potential is inherently constrained by the limited rollout depth.

Overall, these results demonstrate that our EDU framework offers a more robust paradigm than MCTS. 
While P-EDU serves as an effective strategy for resource-constrained scenarios by pruning low-confidence branches, the standard EDU sampling provides the highest performance ceiling.
In contrast, MCTS is limited by its local look-ahead mechanism.
Therefore, the optimal strategy depends on the computational budget: P-EDU for efficiency, and standard EDU for maximizing solution quality.

Furthermore, our EDU framework (including P-EDU) offers a fundamental advantage over MCTS in mitigating the ``cheating'' issue—where high intermediate rewards fail to yield correct final answers. 
Unlike MCTS, which relies heavily on local step scores and may prematurely prune promising branches based on misleading intermediate signals, EDU sampling evaluates the complete solution trajectory. 
By selecting the highest-scoring solution only after the entire generation process is complete, our method effectively bypasses local optima and false high scores, ensuring a more robust alignment between process rewards and final answer correctness.

\begin{figure}[t!]
    \centering
    \includegraphics[width=0.9\linewidth]{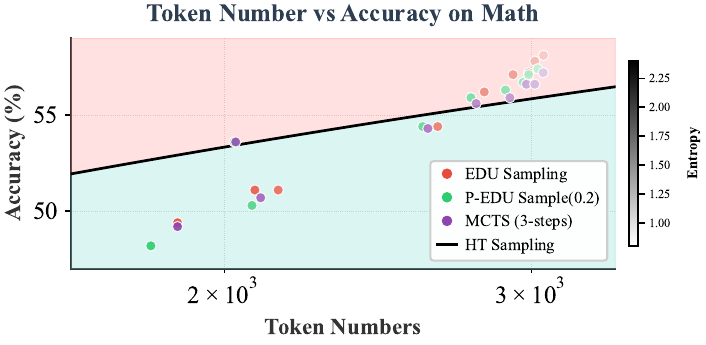}
    \caption{This figure illustrates the relationship between token count and accuracy on the MATH test set under a Max Branch Number of $8$, with the performance of (High-Temperature) HT Sampling across varying token counts fitted as the baseline. On the MATH test set, most data points for both \textcolor{color9}{EDU Sampling} and \textcolor{color10}{P-EDU(0.2) Sampling} lie above this baseline. Notably, as the entropy threshold increases, token counts decrease alongside a corresponding drop in accuracy. Additionally, \textcolor{color11}{MCTS} also surpasses the HT Sampling baseline when the entropy threshold is reduced.}
    \label{fig:entropy_acc_token}
\end{figure}

\subsection{Ablation}

To further investigate the impact of decoding strategies, we introduce a variant called Sample-EDU PRM.
Different from the Greedy-EDU PRM, which utilizes a deterministic greedy decoding approach, Sample-EDU PRM employs stochastic sampling (with temperature $t=0.7$) during the decoding phase whenever no anchor is detected, while keeping all other parameters unchanged, including training methods and the base model.

Our experimental results indicate that Greedy-EDU PRM consistently achieves higher accuracy as the sample size increases (Figure~\ref{fig:comparison_method_of_prm}). 
This improvement can be largely attributed to the deterministic nature of greedy decoding, which helps maintain reasoning consistency throughout the EDU segmentation process.
When combined with entropy-thresholded branching, this method strikes a balance between solution diversity and stability, effectively avoiding the additional noise often associated with stochastic sampling.

In contrast, Sample-EDU leverages stochastic decoding to enhance diversity among candidate solutions. 
However, this increased diversity comes at the cost of greater variability and noise, which tends to weaken the model’s inductive bias and makes performance evaluation less reliable. 
Overall, these findings highlight the trade-offs between diversity and consistency in reasoning, suggesting that a deterministic approach may be better suited for maintaining robust performance in EDU-PRM.

\section{Analysis: Entropy Threshold, Accuracy, and Token Count}\label{sec:entropy}

\paragraph{Definition and Relative Branch Depth}
For a solution trace with \(L\) tokens, let a branch occur at token index \(d\) \((1 \le d \le L)\). We define the relative depth as $r = \frac{d}{L}$.
Aggregating \(r\) across traces into a heat map (Figure~\ref{fig:heatmap_max}) provides a normalized view of where branching tends to concentrate along the trajectory. This metric serves as the foundation for our subsequent analyses on branch timing and behavior.

\subsection{Effect of Entropy Threshold on Branch Timing}
With the relative branch depth metric established, we next examine how the entropy threshold influences the timing of branch points. Figure~\ref{fig:heatmap_entropy} and Table~\ref{tab:math8_performance} and Table~\ref{tab:oly8_performance} show that lowering the entropy threshold shifts branch points earlier in the sequence. A stricter threshold induces earlier branching by pruning diffuse exploratory branches, focusing the search on high-probability paths. Figure~\ref{fig:heatmap_max} further demonstrates that, under selected thresholds, EDU sampling often branches near the very start, resulting in a sharply peaked distribution of relative depths. These results indicate that entropy-based control can effectively modulate when and where branching occurs.

\subsection{Lexical Characteristics of Branch Nodes}
Having identified where branching tends to occur, we now investigate the lexical nature of branch-point tokens. We examine the full-word forms of branch-point tokens and rank words by their branch-point frequency (Figures~\ref{fig:word_frequency_dataset}–\ref{fig:word_frequency_edu}, MATH and OLY test sets). High-frequency items are predominantly function words (e.g., ``then'', ``if'') or light discourse operators (e.g., ``thus'', ``so''). This observation supports our hypothesis that high-entropy tokens act as structural pivots, forming natural boundaries for controlled branching in EDU PRM. The prevalence of such words at branch points suggests that semantic structure guides the branching process.

\subsection{Accuracy–Token Trade-off}
These insights into branch timing and lexical characteristics inform our understanding of the trade-offs involved in branching strategies. Figure~\ref{fig:entropy_acc_token} reports accuracy versus total generated tokens under varying entropy thresholds on MATH (OLY shown in Figure~\ref{fig:entropy_acc_token_oly}). 
As shown in Figure~\ref{fig:entropy_acc_token}, lowering the entropy threshold from $2.4$ to $0.8$ increases accuracy from $49.4\%$ to $58.1\%$, but also raises the average token count from $1,880$ to $3,047$ per sample. This suggests that practitioners must balance accuracy gains against computational overhead when selecting entropy thresholds.
Notably, the EDU sampling begins to outperform the High-Temperature (HT) sampling only when the threshold is sufficiently low to curtail diffuse early exploration. This trade-off highlights the practical importance of threshold selection in balancing computational cost and solution quality.

Furthermore, lowering the entropy threshold tends to produce longer and more detailed reasoning paths, which may improve solution robustness but also increase resource consumption and potentially affect interpretability. Therefore, the optimal threshold may vary depending on the specific application scenario and resource constraints. Future work could explore adaptive or dynamic thresholding strategies to further enhance the efficiency and flexibility of branching methods.

\section{Conclusion}
We propose an entropy-guided sampling method for training process reward models that significantly advances mathematical reasoning.
Our approach consistently outperforms existing baselines and matches the performance of the sota Qwen2.5-Math-PRM with less training data.
Moreover, EDU sampling improves token efficiency in solution generation.
EDU-PRM demonstrates exceptional data efficiency, attaining new state-of-the-art results with minimal training data.
By integrating pruning strategies like P-EDU sampling for cost-effective exploration, our framework provides complementary tools tailored to diverse task demands.
Overall, EDU-PRM establishes a principled methodology that can balance accuracy, efficiency, and search depth in complex reasoning tasks, with promising avenues for future research in scaling to larger datasets, refining intermediate scoring, and developing adaptive generation strategies to extend its applicability across broader domains.



\section*{Limitations}

Although EDU-PRM demonstrates strong performance, several limitations remain. First, the computational cost of entropy calculation during inference adds a slight overhead compared to standard greedy decoding. Second, our experiments focus primarily on mathematical reasoning; the generalizability to other domains (e.g., coding, creative writing) requires further investigation. Finally, while we reduce the need for human annotation, the quality of the PRM still depends on the base model's capability to generate valid reasoning traces.

\bibliography{custom}

@inproceedings{hendrycks2021measuringmathematicalproblemsolving,
  author       = {Dan Hendrycks and
                  Collin Burns and
                  Saurav Kadavath and
                  Akul Arora and
                  Steven Basart and
                  Eric Tang and
                  Dawn Song and
                  Jacob Steinhardt},
  editor       = {Joaquin Vanschoren and
                  Sai{-}Kit Yeung},
  title        = {Measuring Mathematical Problem Solving With the {MATH} Dataset},
  booktitle    = {Proceedings of the Neural Information Processing Systems Track on
                  Datasets and Benchmarks 1, NeurIPS Datasets and Benchmarks 2021, December
                  2021, virtual},
  year         = {2021},
  url          = {https://datasets-benchmarks-proceedings.neurips.cc/paper/2021/hash/be83ab3ecd0db773eb2dc1b0a17836a1-Abstract-round2.html},
  bibsource    = {dblp computer science bibliography, https://dblp.org}
}

@inproceedings{wang2024chainofthoughtreasoningprompting,
  author       = {Xuezhi Wang and
                  Denny Zhou},
  editor       = {Amir Globersons and
                  Lester Mackey and
                  Danielle Belgrave and
                  Angela Fan and
                  Ulrich Paquet and
                  Jakub M. Tomczak and
                  Cheng Zhang},
  title        = {Chain-of-Thought Reasoning Without Prompting},
  booktitle    = {Advances in Neural Information Processing Systems 38: Annual Conference
                  on Neural Information Processing Systems 2024, NeurIPS 2024, Vancouver,
                  BC, Canada, December 10 - 15, 2024},
  year         = {2024},
  url          = {http://papers.nips.cc/paper\_files/paper/2024/hash/7a8e7fd295aa04eac4b470ae27f8785c-Abstract-Conference.html},
  bibsource    = {dblp computer science bibliography, https://dblp.org}
}

@misc{qwen2025qwen25technicalreport,
      title={Qwen2.5 Technical Report}, 
      author={Qwen and : and An Yang and Baosong Yang and Beichen Zhang and Binyuan Hui and Bo Zheng and Bowen Yu and Chengyuan Li and Dayiheng Liu and Fei Huang and et al.},
      year={2025},
      eprint={2412.15115},
      archivePrefix={arXiv},
      primaryClass={cs.CL},
      url={https://arxiv.org/abs/2412.15115}, 
}

@misc{yang2024qwen2technicalreport,
      title={Qwen2 Technical Report}, 
      author={An Yang and Baosong Yang and Binyuan Hui and Bo Zheng and Bowen Yu and Chang Zhou and  et al.},
      year={2024},
      eprint={2407.10671},
      archivePrefix={arXiv},
      primaryClass={cs.CL},
      url={https://arxiv.org/abs/2407.10671}, 
}

@inproceedings{wang2023math,
  author       = {Peiyi Wang and
                  Lei Li and
                  Zhihong Shao and
                  Runxin Xu and
                  Damai Dai and
                  Yifei Li and
                  Deli Chen and
                  Yu Wu and
                  Zhifang Sui},
  editor       = {Lun{-}Wei Ku and
                  Andre Martins and
                  Vivek Srikumar},
  title        = {Math-Shepherd: Verify and Reinforce LLMs Step-by-step without Human
                  Annotations},
  booktitle    = {Proceedings of the 62nd Annual Meeting of the Association for Computational
                  Linguistics (Volume 1: Long Papers), {ACL} 2024, Bangkok, Thailand,
                  August 11-16, 2024},
  pages        = {9426--9439},
  publisher    = {Association for Computational Linguistics},
  year         = {2024},
  url          = {https://doi.org/10.18653/v1/2024.acl-long.510},
  doi          = {10.18653/V1/2024.ACL-LONG.510},
  bibsource    = {dblp computer science bibliography, https://dblp.org}
}

@article{luo2024improve,
  title={Improve mathematical reasoning in language models by automated process supervision},
  author={Luo, Liangchen and Liu, Yinxiao and Liu, Rosanne and Phatale, Samrat and Lara, Harsh and Li, Yunxuan and Shu, Lei and Zhu, Yun and Meng, Lei and Sun, Jiao and others},
  journal={arXiv preprint arXiv:2406.06592},
  volume={2},
  year={2024}
}

@article{yang2024qwen25mathtechnicalreportmathematical,
  title={Qwen2.5-Math Technical Report: Toward Mathematical Expert Model via Self-Improvement}, 
  author={An Yang and Beichen Zhang and Binyuan Hui and Bofei Gao and Bowen Yu and Chengpeng Li and Dayiheng Liu and Jianhong Tu and Jingren Zhou and Junyang Lin and Keming Lu and Mingfeng Xue and Runji Lin and Tianyu Liu and Xingzhang Ren and Zhenru Zhang},
  journal={arXiv preprint arXiv:2409.12122},
  year={2024}
}

@article{yuan2024freeprocessrewardsprocess,
  author       = {Lifan Yuan and
                  Wendi Li and
                  Huayu Chen and
                  Ganqu Cui and
                  Ning Ding and
                  Kai Zhang and
                  Bowen Zhou and
                  Zhiyuan Liu and
                  Hao Peng},
  title        = {Free Process Rewards without Process Labels},
  journal      = {CoRR},
  volume       = {abs/2412.01981},
  year         = {2024},
  url          = {https://doi.org/10.48550/arXiv.2412.01981},
  doi          = {10.48550/ARXIV.2412.01981},
  eprinttype    = {arXiv},
  eprint       = {2412.01981},
  bibsource    = {dblp computer science bibliography, https://dblp.org}
}

@inproceedings{zhang2025lessonsdevelopingprocessreward,
  author       = {Zhenru Zhang and
                  Chujie Zheng and
                  Yangzhen Wu and
                  Beichen Zhang and
                  Runji Lin and
                  Bowen Yu and
                  Dayiheng Liu and
                  Jingren Zhou and
                  Junyang Lin},
  editor       = {Wanxiang Che and
                  Joyce Nabende and
                  Ekaterina Shutova and
                  Mohammad Taher Pilehvar},
  title        = {The Lessons of Developing Process Reward Models in Mathematical Reasoning},
  booktitle    = {Findings of the Association for Computational Linguistics, {ACL} 2025,
                  Vienna, Austria, July 27 - August 1, 2025},
  pages        = {10495--10516},
  publisher    = {Association for Computational Linguistics},
  year         = {2025},
  url          = {https://aclanthology.org/2025.findings-acl.547/},
  bibsource    = {dblp computer science bibliography, https://dblp.org}
}

@inproceedings{lightman2023letsverifystepstep,
  author       = {Hunter Lightman and
                  Vineet Kosaraju and
                  Yuri Burda and
                  Harrison Edwards and
                  Bowen Baker and
                  Teddy Lee and
                  Jan Leike and
                  John Schulman and
                  Ilya Sutskever and
                  Karl Cobbe},
  title        = {Let's Verify Step by Step},
  booktitle    = {The Twelfth International Conference on Learning Representations,
                  {ICLR} 2024, Vienna, Austria, May 7-11, 2024},
  publisher    = {OpenReview.net},
  year         = {2024},
  url          = {https://openreview.net/forum?id=v8L0pN6EOi},
  bibsource    = {dblp computer science bibliography, https://dblp.org}
}

@inproceedings{wang2024mathshepherdverifyreinforcellms,
  author       = {Peiyi Wang and
                  Lei Li and
                  Zhihong Shao and
                  Runxin Xu and
                  Damai Dai and
                  Yifei Li and
                  Deli Chen and
                  Yu Wu and
                  Zhifang Sui},
  editor       = {Lun{-}Wei Ku and
                  Andre Martins and
                  Vivek Srikumar},
  title        = {Math-Shepherd: Verify and Reinforce LLMs Step-by-step without Human
                  Annotations},
  booktitle    = {Proceedings of the 62nd Annual Meeting of the Association for Computational
                  Linguistics (Volume 1: Long Papers), {ACL} 2024, Bangkok, Thailand,
                  August 11-16, 2024},
  pages        = {9426--9439},
  publisher    = {Association for Computational Linguistics},
  year         = {2024},
  url          = {https://doi.org/10.18653/v1/2024.acl-long.510},
  doi          = {10.18653/V1/2024.ACL-LONG.510},
  bibsource    = {dblp computer science bibliography, https://dblp.org}
}

@inproceedings{wang2023selfconsistencyimproveschainthought,
  author       = {Xuezhi Wang and
                  Jason Wei and
                  Dale Schuurmans and
                  Quoc V. Le and
                  Ed H. Chi and
                  Sharan Narang and
                  Aakanksha Chowdhery and
                  Denny Zhou},
  title        = {Self-Consistency Improves Chain of Thought Reasoning in Language Models},
  booktitle    = {The Eleventh International Conference on Learning Representations,
                  {ICLR} 2023, Kigali, Rwanda, May 1-5, 2023},
  publisher    = {OpenReview.net},
  year         = {2023},
  url          = {https://openreview.net/forum?id=1PL1NIMMrw},
  bibsource    = {dblp computer science bibliography, https://dblp.org}
}

@article{luo2024improvemathematicalreasoninglanguage,
  author       = {Liangchen Luo and
                  Yinxiao Liu and
                  Rosanne Liu and
                  Samrat Phatale and
                  Harsh Lara and
                  Yunxuan Li and
                  Lei Shu and
                  Yun Zhu and
                  Lei Meng and
                  Jiao Sun and
                  Abhinav Rastogi},
  title        = {Improve Mathematical Reasoning in Language Models by Automated Process
                  Supervision},
  journal      = {CoRR},
  volume       = {abs/2406.06592},
  year         = {2024},
  url          = {https://doi.org/10.48550/arXiv.2406.06592},
  doi          = {10.48550/ARXIV.2406.06592},
  eprinttype    = {arXiv},
  eprint       = {2406.06592},
  bibsource    = {dblp computer science bibliography, https://dblp.org}
}

@inproceedings{kwon2023efficientmemorymanagementlarge,
  author       = {Woosuk Kwon and
                  Zhuohan Li and
                  Siyuan Zhuang and
                  Ying Sheng and
                  Lianmin Zheng and
                  Cody Hao Yu and
                  Joseph Gonzalez and
                  Hao Zhang and
                  Ion Stoica},
  editor       = {Jason Flinn and
                  Margo I. Seltzer and
                  Peter Druschel and
                  Antoine Kaufmann and
                  Jonathan Mace},
  title        = {Efficient Memory Management for Large Language Model Serving with
                  PagedAttention},
  booktitle    = {Proceedings of the 29th Symposium on Operating Systems Principles,
                  {SOSP} 2023, Koblenz, Germany, October 23-26, 2023},
  pages        = {611--626},
  publisher    = {{ACM}},
  year         = {2023},
  url          = {https://doi.org/10.1145/3600006.3613165},
  doi          = {10.1145/3600006.3613165},
  bibsource    = {dblp computer science bibliography, https://dblp.org}
}

@inproceedings{aminabadi2022deepspeedinferenceenablingefficient,
  author       = {Reza Yazdani Aminabadi and
                  Samyam Rajbhandari and
                  Ammar Ahmad Awan and
                  Cheng Li and
                  Du Li and
                  Elton Zheng and
                  Olatunji Ruwase and
                  Shaden Smith and
                  Minjia Zhang and
                  Jeff Rasley and
                  Yuxiong He},
  editor       = {Felix Wolf and
                  Sameer Shende and
                  Candace Culhane and
                  Sadaf R. Alam and
                  Heike Jagode},
  title        = {DeepSpeed- Inference: Enabling Efficient Inference of Transformer
                  Models at Unprecedented Scale},
  booktitle    = {{SC22:} International Conference for High Performance Computing, Networking,
                  Storage and Analysis, Dallas, TX, USA, November 13-18, 2022},
  pages        = {46:1--46:15},
  publisher    = {{IEEE}},
  year         = {2022},
  url          = {https://doi.org/10.1109/SC41404.2022.00051},
  doi          = {10.1109/SC41404.2022.00051},
  bibsource    = {dblp computer science bibliography, https://dblp.org}
}

@inproceedings{mu2024rulebasedrewardslanguage,
  author       = {Tong Mu and
                  Alec Helyar and
                  Johannes Heidecke and
                  Joshua Achiam and
                  Andrea Vallone and
                  Ian Kivlichan and
                  Molly Lin and
                  Alex Beutel and
                  John Schulman and
                  Lilian Weng},
  editor       = {Amir Globersons and
                  Lester Mackey and
                  Danielle Belgrave and
                  Angela Fan and
                  Ulrich Paquet and
                  Jakub M. Tomczak and
                  Cheng Zhang},
  title        = {Rule Based Rewards for Language Model Safety},
  booktitle    = {Advances in Neural Information Processing Systems 38: Annual Conference
                  on Neural Information Processing Systems 2024, NeurIPS 2024, Vancouver,
                  BC, Canada, December 10 - 15, 2024},
  year         = {2024},
  url          = {http://papers.nips.cc/paper\_files/paper/2024/hash/c4e380fb74dec9da9c7212e834657aa9-Abstract-Conference.html},
  bibsource    = {dblp computer science bibliography, https://dblp.org}
}

@inproceedings{zheng2023judgingllmasajudgemtbenchchatbot,
  author       = {Lianmin Zheng and
                  Wei{-}Lin Chiang and
                  Ying Sheng and
                  Siyuan Zhuang and
                  Zhanghao Wu and
                  Yonghao Zhuang and
                  Zi Lin and
                  Zhuohan Li and
                  Dacheng Li and
                  Eric P. Xing and
                  Hao Zhang and
                  Joseph E. Gonzalez and
                  Ion Stoica},
  editor       = {Alice Oh and
                  Tristan Naumann and
                  Amir Globerson and
                  Kate Saenko and
                  Moritz Hardt and
                  Sergey Levine},
  title        = {Judging LLM-as-a-Judge with MT-Bench and Chatbot Arena},
  booktitle    = {Advances in Neural Information Processing Systems 36: Annual Conference
                  on Neural Information Processing Systems 2023, NeurIPS 2023, New Orleans,
                  LA, USA, December 10 - 16, 2023},
  year         = {2023},
  url          = {http://papers.nips.cc/paper\_files/paper/2023/hash/91f18a1287b398d378ef22505bf41832-Abstract-Datasets\_and\_Benchmarks.html},
  bibsource    = {dblp computer science bibliography, https://dblp.org}
}

@inproceedings{wang2024interpretablepreferencesmultiobjectivereward,
  author       = {Haoxiang Wang and
                  Wei Xiong and
                  Tengyang Xie and
                  Han Zhao and
                  Tong Zhang},
  editor       = {Yaser Al{-}Onaizan and
                  Mohit Bansal and
                  Yun{-}Nung Chen},
  title        = {Interpretable Preferences via Multi-Objective Reward Modeling and
                  Mixture-of-Experts},
  booktitle    = {Findings of the Association for Computational Linguistics: {EMNLP}
                  2024, Miami, Florida, USA, November 12-16, 2024},
  pages        = {10582--10592},
  publisher    = {Association for Computational Linguistics},
  year         = {2024},
  url          = {https://doi.org/10.18653/v1/2024.findings-emnlp.620},
  doi          = {10.18653/V1/2024.FINDINGS-EMNLP.620},
  bibsource    = {dblp computer science bibliography, https://dblp.org}
}

@misc{deepseekai2025deepseekr1incentivizingreasoningcapability,
      title={DeepSeek-R1: Incentivizing Reasoning Capability in LLMs via Reinforcement Learning}, 
      author={DeepSeek-AI and Daya Guo and Dejian Yang and Haowei Zhang and Junxiao Song and Ruoyu Zhang and Runxin Xu and et al.},
      year={2025},
      eprint={2501.12948},
      archivePrefix={arXiv},
      primaryClass={cs.CL},
      url={https://arxiv.org/abs/2501.12948}, 
}

@article{wu2024metarewardinglanguagemodelsselfimproving,
  author       = {Tianhao Wu and
                  Weizhe Yuan and
                  Olga Golovneva and
                  Jing Xu and
                  Yuandong Tian and
                  Jiantao Jiao and
                  Jason Weston and
                  Sainbayar Sukhbaatar},
  title        = {Meta-Rewarding Language Models: Self-Improving Alignment with LLM-as-a-Meta-Judge},
  journal      = {CoRR},
  volume       = {abs/2407.19594},
  year         = {2024},
  url          = {https://doi.org/10.48550/arXiv.2407.19594},
  doi          = {10.48550/ARXIV.2407.19594},
  eprinttype    = {arXiv},
  eprint       = {2407.19594},
  bibsource    = {dblp computer science bibliography, https://dblp.org}
}

@misc{katzgraber2011introductionmontecarlomethods,
      title={Introduction to Monte Carlo Methods}, 
      author={Helmut G. Katzgraber},
      year={2011},
      eprint={0905.1629},
      archivePrefix={arXiv},
      primaryClass={cond-mat.stat-mech},
      url={https://arxiv.org/abs/0905.1629}, 
}

@misc{openai2024gpt4ocard,
      title={GPT-4o System Card}, 
      author={OpenAI and : and Aaron Hurst and Adam Lerer and Adam P. Goucher and Adam Perelman and Aditya Ramesh and Aidan Clark and AJ Ostrow and Akila Welihinda and Alan Hayes and et al.},
      year={2024},
      eprint={2410.21276},
      archivePrefix={arXiv},
      primaryClass={cs.CL},
      url={https://arxiv.org/abs/2410.21276}, 
}

@article{deepseekai2025deepseekv3technicalreport,
  author       = {DeepSeek{-}AI and
                  Aixin Liu and
                  Bei Feng and
                  Bing Xue and
                  Bingxuan Wang and
                  Bochao Wu and
                  Chengda Lu and
                  Chenggang Zhao and
                  Chengqi Deng and
                  Chenyu Zhang and
                  Chong Ruan and
                  Damai Dai and
                  Daya Guo and et al.},
  title        = {DeepSeek-V3 Technical Report},
  journal      = {CoRR},
  volume       = {abs/2412.19437},
  year         = {2024},
  url          = {https://doi.org/10.48550/arXiv.2412.19437},
  doi          = {10.48550/ARXIV.2412.19437},
  eprinttype    = {arXiv},
  eprint       = {2412.19437},
  bibsource    = {dblp computer science bibliography, https://dblp.org}
}

@article{yang2025qwen3technicalreport,
  author       = {An Yang and
                  Anfeng Li and
                  Baosong Yang and
                  Beichen Zhang and
                  Binyuan Hui and
                  Bo Zheng and
                  Bowen Yu and
                  Chang Gao and et al.},
  title        = {Qwen3 Technical Report},
  journal      = {CoRR},
  volume       = {abs/2505.09388},
  year         = {2025},
  url          = {https://doi.org/10.48550/arXiv.2505.09388},
  doi          = {10.48550/ARXIV.2505.09388},
  eprinttype    = {arXiv},
  eprint       = {2505.09388},
  bibsource    = {dblp computer science bibliography, https://dblp.org}
}

@inproceedings{wei2023chainofthoughtpromptingelicitsreasoning,
  author       = {Jason Wei and
                  Xuezhi Wang and
                  Dale Schuurmans and
                  Maarten Bosma and
                  Brian Ichter and
                  Fei Xia and
                  Ed H. Chi and
                  Quoc V. Le and
                  Denny Zhou},
  editor       = {Sanmi Koyejo and
                  S. Mohamed and
                  A. Agarwal and
                  Danielle Belgrave and
                  K. Cho and
                  A. Oh},
  title        = {Chain-of-Thought Prompting Elicits Reasoning in Large Language Models},
  booktitle    = {Advances in Neural Information Processing Systems 35: Annual Conference
                  on Neural Information Processing Systems 2022, NeurIPS 2022, New Orleans,
                  LA, USA, November 28 - December 9, 2022},
  year         = {2022},
  url          = {http://papers.nips.cc/paper\_files/paper/2022/hash/9d5609613524ecf4f15af0f7b31abca4-Abstract-Conference.html},
  bibsource    = {dblp computer science bibliography, https://dblp.org}
}

@article{murphy2025reinforcementlearningoverview,
  author       = {Kevin Murphy},
  title        = {Reinforcement Learning: An Overview},
  journal      = {CoRR},
  volume       = {abs/2412.05265},
  year         = {2024},
  url          = {https://doi.org/10.48550/arXiv.2412.05265},
  doi          = {10.48550/ARXIV.2412.05265},
  eprinttype    = {arXiv},
  eprint       = {2412.05265},
  bibsource    = {dblp computer science bibliography, https://dblp.org}
}

@inproceedings{he2024olympiadbenchchallengingbenchmarkpromoting,
  author       = {Chaoqun He and
                  Renjie Luo and
                  Yuzhuo Bai and
                  Shengding Hu and
                  Zhen Leng Thai and
                  Junhao Shen and
                  Jinyi Hu and
                  Xu Han and
                  Yujie Huang and
                  Yuxiang Zhang and
                  Jie Liu and
                  Lei Qi and
                  Zhiyuan Liu and
                  Maosong Sun},
  editor       = {Lun{-}Wei Ku and
                  Andre Martins and
                  Vivek Srikumar},
  title        = {OlympiadBench: {A} Challenging Benchmark for Promoting {AGI} with
                  Olympiad-Level Bilingual Multimodal Scientific Problems},
  booktitle    = {Proceedings of the 62nd Annual Meeting of the Association for Computational
                  Linguistics (Volume 1: Long Papers), {ACL} 2024, Bangkok, Thailand,
                  August 11-16, 2024},
  pages        = {3828--3850},
  publisher    = {Association for Computational Linguistics},
  year         = {2024},
  url          = {https://doi.org/10.18653/v1/2024.acl-long.211},
  doi          = {10.18653/V1/2024.ACL-LONG.211},
  bibsource    = {dblp computer science bibliography, https://dblp.org}
}

@article{cobbe2021trainingverifierssolvemath,
  author       = {Karl Cobbe and
                  Vineet Kosaraju and
                  Mohammad Bavarian and
                  Mark Chen and
                  Heewoo Jun and
                  Lukasz Kaiser and
                  Matthias Plappert and
                  Jerry Tworek and
                  Jacob Hilton and
                  Reiichiro Nakano and
                  Christopher Hesse and
                  John Schulman},
  title        = {Training Verifiers to Solve Math Word Problems},
  journal      = {CoRR},
  volume       = {abs/2110.14168},
  year         = {2021},
  url          = {https://arxiv.org/abs/2110.14168},
  eprinttype    = {arXiv},
  eprint       = {2110.14168},
  bibsource    = {dblp computer science bibliography, https://dblp.org}
}

@article{xie2024montecarlotreesearch,
  author       = {Yuxi Xie and
                  Anirudh Goyal and
                  Wenyue Zheng and
                  Min{-}Yen Kan and
                  Timothy P. Lillicrap and
                  Kenji Kawaguchi and
                  Michael Shieh},
  title        = {Monte Carlo Tree Search Boosts Reasoning via Iterative Preference
                  Learning},
  journal      = {CoRR},
  volume       = {abs/2405.00451},
  year         = {2024},
  url          = {https://doi.org/10.48550/arXiv.2405.00451},
  doi          = {10.48550/ARXIV.2405.00451},
  eprinttype    = {arXiv},
  eprint       = {2405.00451},
  bibsource    = {dblp computer science bibliography, https://dblp.org}
}

@inproceedings{zhang2024restmctsllmselftrainingprocess,
  author       = {Dan Zhang and
                  Sining Zhoubian and
                  Ziniu Hu and
                  Yisong Yue and
                  Yuxiao Dong and
                  Jie Tang},
  editor       = {Amir Globersons and
                  Lester Mackey and
                  Danielle Belgrave and
                  Angela Fan and
                  Ulrich Paquet and
                  Jakub M. Tomczak and
                  Cheng Zhang},
  title        = {ReST-MCTS*: {LLM} Self-Training via Process Reward Guided Tree Search},
  booktitle    = {Advances in Neural Information Processing Systems 38: Annual Conference
                  on Neural Information Processing Systems 2024, NeurIPS 2024, Vancouver,
                  BC, Canada, December 10 - 15, 2024},
  year         = {2024},
  url          = {http://papers.nips.cc/paper\_files/paper/2024/hash/76ec4dc30e9faaf0e4b6093eaa377218-Abstract-Conference.html},
  bibsource    = {dblp computer science bibliography, https://dblp.org}
}

@inproceedings{zheng2025processbenchidentifyingprocesserrors,
  author       = {Chujie Zheng and
                  Zhenru Zhang and
                  Beichen Zhang and
                  Runji Lin and
                  Keming Lu and
                  Bowen Yu and
                  Dayiheng Liu and
                  Jingren Zhou and
                  Junyang Lin},
  editor       = {Wanxiang Che and
                  Joyce Nabende and
                  Ekaterina Shutova and
                  Mohammad Taher Pilehvar},
  title        = {ProcessBench: Identifying Process Errors in Mathematical Reasoning},
  booktitle    = {Proceedings of the 63rd Annual Meeting of the Association for Computational
                  Linguistics (Volume 1: Long Papers), {ACL} 2025, Vienna, Austria,
                  July 27 - August 1, 2025},
  pages        = {1009--1024},
  publisher    = {Association for Computational Linguistics},
  year         = {2025},
  url          = {https://aclanthology.org/2025.acl-long.50/},
  bibsource    = {dblp computer science bibliography, https://dblp.org}
}

@inproceedings{tang2024mathscalescalinginstructiontuning,
  author       = {Zhengyang Tang and
                  Xingxing Zhang and
                  Benyou Wang and
                  Furu Wei},
  title        = {MathScale: Scaling Instruction Tuning for Mathematical Reasoning},
  booktitle    = {Forty-first International Conference on Machine Learning, {ICML} 2024,
                  Vienna, Austria, July 21-27, 2024},
  publisher    = {OpenReview.net},
  year         = {2024},
  url          = {https://openreview.net/forum?id=Kjww7ZN47M},
  bibsource    = {dblp computer science bibliography, https://dblp.org}
}

@inproceedings{sun2024easytohardgeneralizationscalablealignment,
  author       = {Zhiqing Sun and
                  Longhui Yu and
                  Yikang Shen and
                  Weiyang Liu and
                  Yiming Yang and
                  Sean Welleck and
                  Chuang Gan},
  editor       = {Amir Globersons and
                  Lester Mackey and
                  Danielle Belgrave and
                  Angela Fan and
                  Ulrich Paquet and
                  Jakub M. Tomczak and
                  Cheng Zhang},
  title        = {Easy-to-Hard Generalization: Scalable Alignment Beyond Human Supervision},
  booktitle    = {Advances in Neural Information Processing Systems 38: Annual Conference
                  on Neural Information Processing Systems 2024, NeurIPS 2024, Vancouver,
                  BC, Canada, December 10 - 15, 2024},
  year         = {2024},
  url          = {http://papers.nips.cc/paper\_files/paper/2024/hash/5b6346a05a537d4cdb2f50323452a9fe-Abstract-Conference.html},
  bibsource    = {dblp computer science bibliography, https://dblp.org}
}

@inproceedings{yin2025dynamicgeneralizableprocessreward,
  author       = {Zhangyue Yin and
                  Qiushi Sun and
                  Zhiyuan Zeng and
                  Qinyuan Cheng and
                  Xipeng Qiu and
                  Xuanjing Huang},
  editor       = {Wanxiang Che and
                  Joyce Nabende and
                  Ekaterina Shutova and
                  Mohammad Taher Pilehvar},
  title        = {Dynamic and Generalizable Process Reward Modeling},
  booktitle    = {Proceedings of the 63rd Annual Meeting of the Association for Computational
                  Linguistics (Volume 1: Long Papers), {ACL} 2025, Vienna, Austria,
                  July 27 - August 1, 2025},
  pages        = {4203--4233},
  publisher    = {Association for Computational Linguistics},
  year         = {2025},
  url          = {https://aclanthology.org/2025.acl-long.212/},
  bibsource    = {dblp computer science bibliography, https://dblp.org}
}

@misc{cheng2025reasoningexplorationentropyperspective,
      title={Reasoning with Exploration: An Entropy Perspective on Reinforcement Learning for LLMs}, 
      author={Daixuan Cheng and Shaohan Huang and Xuekai Zhu and Bo Dai and Wayne Xin Zhao and Zhenliang Zhang and Furu Wei},
      year={2025},
      eprint={2506.14758},
      archivePrefix={arXiv},
      primaryClass={cs.CL},
      url={https://arxiv.org/abs/2506.14758}, 
}

@misc{zhang2025entropyregularizedprocessrewardmodel,
      title={Entropy-Regularized Process Reward Model}, 
      author={Hanning Zhang and Pengcheng Wang and Shizhe Diao and Yong Lin and Rui Pan and Hanze Dong and Dylan Zhang and Pavlo Molchanov and Tong Zhang},
      year={2025},
      eprint={2412.11006},
      archivePrefix={arXiv},
      primaryClass={cs.LG},
      url={https://arxiv.org/abs/2412.11006}, 
}

@misc{ye2025uncertaintyawarestepwiseverificationgenerative,
      title={Uncertainty-Aware Step-wise Verification with Generative Reward Models}, 
      author={Zihuiwen Ye and Luckeciano Carvalho Melo and Younesse Kaddar and Phil Blunsom and Sam Staton and Yarin Gal},
      year={2025},
      eprint={2502.11250},
      archivePrefix={arXiv},
      primaryClass={cs.CL},
      url={https://arxiv.org/abs/2502.11250}, 
}

@misc{han2025uncertaintybasedmethodsautomatedprocess,
      title={Uncertainty-Based Methods for Automated Process Reward Data Construction and Output Aggregation in Mathematical Reasoning}, 
      author={Jiuzhou Han and Wray Buntine and Ehsan Shareghi},
      year={2025},
      eprint={2508.01773},
      archivePrefix={arXiv},
      primaryClass={cs.AI},
      url={https://arxiv.org/abs/2508.01773}, 
}

\appendix

\section{Appendix}
\label{sec:appendix}

\subsection{The use of Large Language Models}
Large Language Models (LLMs) were used in this work solely as writing assistance tools. Specifically LLMs were employed to check for spelling errors, grammatical mistakes, and to improve the fluency and precision of expression in the paper. The LLMs did not contribute to research methodology experimental design, or data analysis. All scientific content, ideas, and conclusions presented in this paper are entirely the authors’ own work.

\subsection{ProcessBench}\label{app:process_bench}

Table~\ref{tab:model_performance} provides a comprehensive comparison of various PRM models, including Math-Shepherd, Omega, EDU variants, and Qwen-series, across three ProcessBench subsets: GSM8K, MATH, and OlympiaBench. For each dataset, we report results for both 7B and 72B model scales, including accuracy, F1 score, precision, and recall. The best performance for each metric is highlighted in bold. This detailed breakdown enables a more granular understanding of each model's strengths and limitations across different reasoning benchmarks and evaluation metrics.

\begin{table*}[h]
\centering
\begin{tabular}{llcccc}
\toprule
\multicolumn{2}{l}{\textbf{Task}} & \textbf{Accuracy} & \textbf{F1} & \textbf{Precision} & \textbf{Recall} \\
\midrule
\multicolumn{6}{l}{\textbf{GSM8K}} \\
\midrule
\multirow{5}{*}{7B} & Math-Shepherd PRM & 57.2 & 0.682 & 0.545 & 0.91 \\
& Omega PRM & 57.5 & 0.31 & 0.844 & 0.19 \\
& Sample EDU PRM & 52.5 & 0.677 & 0.513 & \textbf{0.995} \\
& Greedy EDU PRM & 55.2 & 0.218 & \textbf{0.862} & 0.125 \\
& 
Qwen2.5-Math-PRM-7B & \textbf{88.8} & \textbf{0.895} & 0.838 & 0.96 \\
\midrule
\multirow{5}{*}{72B} & Math-Shepherd PRM & 74.5 & 0.803 & 0.671 & \textbf{1} \\
& Omega PRM & 90.5 & 0.908 & 0.882 & 0.935 \\
& Sample EDU PRM & 71 & 0.778 & 0.637 & \textbf{1} \\
& Greedy EDU PRM & 94.2 & 0.95 & 0.909 & 0.995 \\
& 
Qwen2.5-Math-PRM-72B & \textbf{96} & \textbf{0.961} & \textbf{0.938} & 0.985 \\
\midrule
\multicolumn{6}{l}{\textbf{MATH}} \\
\midrule
\multirow{5}{*}{7B} & Math-Shepherd PRM & 62.9 & 0.659 & 0.615 & 0.71 \\
& Omega PRM & 58 & 0.295 & 0.917 & 0.176 \\
& Sample EDU PRM & 59.2 & 0.689 & 0.559 & \textbf{0.898} \\
& Greedy EDU PRM & 56.2 & 0.229 & \textbf{0.956} & 0.13 \\
& 
Qwen2.5-Math-PRM-7B & \textbf{82.4} & \textbf{0.82} & 0.839 & 0.802 \\
\midrule
\multirow{5}{*}{72B} & Math-Shepherd PRM & 77.8 & 0.805 & 0.727 & 0.902 \\
& Omega PRM & 79.8 & 0.763 & \textbf{0.923} & 0.65 \\
& Sample EDU PRM & 76.4 & 0.795 & 0.709 & \textbf{0.906} \\
& Greedy EDU PRM & \textbf{88.4} & \textbf{0.882} & 0.904 & 0.862 \\
& 
Qwen2.5-Math-PRM-72B & 87.8 & 0.872 & 0.918 & 0.83 \\
\midrule
\multicolumn{6}{l}{\textbf{OlympiaBench}} \\
\midrule
\multirow{5}{*}{7B} & Math-Shepherd PRM & 53.6 & 0.539 & 0.541 & 0.536 \\
& Omega PRM & 51.3 & 0.079 & 0.724 & 0.042 \\
& Sample EDU PRM & 53.8 & 0.636 & 0.528 & \textbf{0.798} \\
& Greedy EDU PRM & 51.7 & 0.083 & \textbf{0.815} & 0.004 \\
& 
Qwen2.5-Math-PRM-7B & \textbf{74.1} & \textbf{0.721} & 0.785 & 0.666 \\
\midrule
\multirow{5}{*}{72B} & Math-Shepherd PRM & 71 & 0.74 & 0.691 & \textbf{0.796} \\
& Omega PRM & 66.1 & 0.553 & 0.816 & 0.418 \\
& Sample EDU PRM & 69.7 & 0.723 & 0.67 & 0.786 \\
& Greedy EDU PRM & 77.2 & 0.762 & 0.801 & 0.726 \\
& 
Qwen2.5-Math-PRM-72B & \textbf{79.8} & \textbf{0.779} & \textbf{0.86} & 0.712 \\
\bottomrule
\end{tabular}
\caption{Performance comparison of different PRM models (Math-Shepherd, Omega, EDU, Qwen-series)  on three ProcessBench subsets: GSM8K, MATH, and OLY. For each dataset, results are reported for both 7B and 72B model sizes, including metrics for accuracy, F1 score, precision, and recall. The best results for each metric are highlighted in bold.}
\label{tab:model_performance}
\end{table*}

\subsection{Experimental Environment, Training Configuration and Dataset Details}\label{app:dataset}

This appendix provides detailed information on the experimental platform, framework selection, model training settings, and evaluation datasets used in this study, ensuring the reproducibility of the experiments.

\subsubsection{Experimental Platform and Framework}
All experiments were conducted on the \textbf{Ascend 910B platform} to ensure stable computing performance. Different frameworks were adopted for specific experimental phases to optimize efficiency:

\begin{itemize}
    \item \textbf{PRM Training Data Production}: Employed the DeepSpeed inference framework to accelerate data processing and generation.
    \item \textbf{Solution Generation Phase}: Utilized the VLLM inference framework, which is optimized for high-throughput and low-latency text generation tasks.
    \item \textbf{PRM Training}: Adopted the Mindspeed framework, selected for its efficiency in training large-scale models for preference learning.
\end{itemize}

\subsubsection{Model Training Configuration}
Comparative experiments were conducted on two base models with different parameter scales (7B and 72B), using identical training configurations to ensure result consistency and comparability:
\begin{enumerate}
    \item Initial learning rate: $10^{-6}$
    \item Minimum learning rate (lower bound): $10^{-7}$
    \item Warmup mechanism: Applied with a warmup ratio of 0.01 to stabilize parameter updates in the early training stage.
    \item Cosine Annealing: Adopted a cosine strategy for subsequent learning rate adjustment, balancing late-stage convergence and overfitting prevention.
    \item \textbf{Training Cycle and Checkpoint Management}:
        \begin{itemize}
            \item Total training epochs: 5 (uniformly set for both models).
            \item Checkpoint (ckpt) saving: Automatically saved at the end of each epoch to facilitate subsequent result screening and experiment reproducibility.
            \item Optimal Checkpoint Selection: Compared the core metrics (e.g., accuracy, perplexity) of checkpoints from 5 epochs on the validation set; the checkpoint with the best performance was selected as the basis for final result reporting, ensuring objectivity and representativeness.
        \end{itemize}
\end{enumerate}

\subsubsection{Details of Evaluation Datasets}
Five datasets covering different difficulty levels (from elementary to university-level) and task types (math reasoning, multi-step problem-solving) were used to comprehensively evaluate the model's generalization and reasoning abilities. The key details of each dataset are presented in Table~\ref{tab:evaluation_datasetss}.

\begin{table*}[ht]
    \centering
    \resizebox{0.9\textwidth}{!}{
    \begin{tabular}{p{3cm}p{6cm}p{5cm}}
        \toprule
        Dataset & Description & Usage in Evaluation \\
        \midrule
        OlympiadBench & Bilingual, multimodal dataset with 8,952 math/physics questions (from Olympiads, college entrance exams); subset ``OE\_TO\_maths\_en\_COMP'' contains 675 problems. & Used the ``OE\_TO\_maths\_en\_COMP'' subset (675 problems) to evaluate the model's performance on competitive/advanced math tasks. \\
        \midrule
        GSM8K & 8,500+ grade school math word problems (linguistically diverse, requiring 2–8 steps of basic arithmetic reasoning); solutions in natural language; 1,319 test data points. & Used 1,319 test data points to evaluate the model's elementary mathematical reasoning and multi-step natural language-based problem-solving skills. \\
        \midrule
        MATH & Consists of 12,500 challenging competition-level mathematics problems, each with detailed step-by-step solutions. We selected 5,000 problems as our test set to evaluate the model’s abilities in complex mathematical reasoning, solution derivation , and answer generation. The MATH dataset serves as a rigorous benchmark for assessing advanced mathematical problem-solving skills. & Used the selected 5,000-test-sample subset to systematically evaluate the model’s reasoning process, step-by-step solution generation, and overall accuracy on advanced math problems. \\
        \midrule
        CollegeMath & $\sim$1100 university-level math problems (covering 6 college math areas; 20\% with images). & Used all test data to assess the model's proficiency in complex, advanced mathematical concepts (relevant to industry and higher education scenarios). \\
        \midrule
        ProcessBench & Three selected subsets: MATH (1,000 samples), OlympiaBench (1,000 samples), GSM8K (400 samples); each sample includes step-by-step error position annotations and final solution correctness labels; balanced positive/negative samples in each subset. & Used to evaluate the model's overall solution correctness. \\
        \bottomrule
    \end{tabular}
    }
    \caption{Key details of evaluation datasets used in our experiments.}
    \label{tab:evaluation_datasetss}
\end{table*}

\subsection{EDU Sampling Whitelist}\label{app:mathcal_s}
\begin{tcolorbox}[breakable]
\textbackslash, \$, \textbackslash n, \textbackslash r, \textvisiblespace, \_, \textvisiblespace\textvisiblespace, :, \textbackslash(, \textbackslash[, \textbackslash\{, \textvisiblespace, \textbackslash], \textbackslash), \textbackslash\}, \textbackslash[, \textbackslash(, (, \textbackslash[, [, \textbackslash], \textbackslash\{, \{, \textbackslash\}, \}
\end{tcolorbox}

\subsection{Evaluation Prompt}\label{app:evaluation_prompt}
We use the following prompt to evaluate the solution, with Qwen3-32B-instruct~\citep{yang2025qwen3technicalreport} as the underlying model. For each test instance, the model is provided with the problem statement and instructed to generate a step-by-step solution. The prompt is designed to encourage detailed reasoning and explicit justification at each step, ensuring the model's output is both accurate and interpretable.

\begin{tcolorbox}[breakable]
  
You are a high school math teacher who is correcting homework. You need to check whether the result from Response is consistent with the Answer. If the result is consistent, you should reply Yes, otherwise you should reply No. You only need to compare the result from Response to the Answer. If the content in the Response does not give the final result, answer No.

Here are some examples

\textbf{[Question]}

If $x = 3$ and $y = 2$, then what is the value of $\dfrac{2x^3-3y^2}{6}$ ?

\textbf{[Response]}

To find the value of the expression $\dfrac{2x^3-3y^2}{6}$ given $x = 3$ and $y = 2$, we substitute the values of $x$ and $y$ into the expression and simplify step by step.

\begin{enumerate}
  \item \textbf{Substitute $x = 3$ and $y = 2$:}
  \[
    \frac{2(3)^3 - 3(2)^2}{6}
  \]

  \item \textbf{Simplify the expression inside the numerator:}
  \[
    3^3 = 27,\qquad 2^2 = 4
  \]
  So,
  \[
    \frac{2(27) - 3(4)}{6}
  \]

  \item \textbf{Further simplify:}
  \[
    2 \times 27 = 54,\qquad 3 \times 4 = 12
  \]
  Hence
  \[
    \frac{54 - 12}{6}
  \]

  \item \textbf{Subtract in the numerator:}
  \[
    54 - 12 = 42 \Rightarrow \frac{42}{6}
  \]

  \item \textbf{Simplify the fraction:}
  \[
    42 \div 6 = 7
  \]
  Therefore the value is $\boxed{7}$.
\end{enumerate}

\textbf{[Answer]}

7

\textbf{[Evaluation]}

Yes

\textbf{[Question]}

Find $\cos^{-1}\!\bigl(\tfrac{1}{2}\bigr)$. Express your answer in radians.

\textbf{[Response]}

To find $\cos^{-1}\!\bigl(\tfrac{1}{2}\bigr)$, we need the angle whose cosine is $\tfrac{1}{2}$. From the $30$-$60$-$90$ triangle, $\cos\bigl(\tfrac{\pi}{6}\bigr)=\tfrac{1}{2}$. Therefore $\cos^{-1}\!\bigl(\tfrac{1}{2}\bigr)=\tfrac{\pi}{6}$. So the answer is $\boxed{\tfrac{\pi}{6}}$.

\textbf{[Answer]}

$\dfrac{\pi}{3}$

\textbf{[Evaluation]}

No

\textbf{[Question]}

Consider two lines: line $l$ parameterized as
\[
x = 1 + 4t,\qquad y = 4 + 3t
\]
and the line $m$ parameterized as
\[
x = -5 + 4s,\qquad y = 6 + 3s.
\]
Let $A$ be a point on line $l$, $B$ be a point on line $m$, and let $P$ be the foot of the perpendicular from $A$ to line $m$.

Then $\overrightarrow{BP}$ is the projection of $\overrightarrow{BA}$ onto some vector $\begin{pmatrix} v_1\\ v_2\end{pmatrix}$ such that $v_1+v_2=-7$. Find $\begin{pmatrix} v_1\\ v_2\end{pmatrix}$.

\textbf{[Response]}

(Working leading to)
\[
\boxed{\begin{pmatrix}-4\\ -3\end{pmatrix}}
\]

\textbf{[Answer]}

$\begin{pmatrix}-4\\ -3\end{pmatrix}$

\textbf{[Evaluation]}

Yes

\textbf{[Question]}

Consider two lines: line $l$ parameterized as
\[
x = 1 + 4t,\qquad y = 4 + 3t
\]
and the line $m$ parameterized as
\[
x = -5 + 4s,\qquad y = 6 + 3s.
\]
Let $A$ be a point on line $l$, $B$ be a point on line $m$, and let $P$ be the foot of the perpendicular from $A$ to line $m$.

Then $\overrightarrow{BP}$ is the projection of $\overrightarrow{BA}$ onto some vector $\begin{pmatrix} v_1\\ v_2\end{pmatrix}$ such that $v_1+v_2=-7$. Find $\begin{pmatrix} v_1\\ v_2\end{pmatrix}$.

\textbf{[Response]}

(An unrelated distance-to-plane calculation producing $4$.)

\textbf{[Answer]}

$\dfrac{10}{3}$

\textbf{[Evaluation]}

No

Note: You only need to compare the result from Response to the Answer.

\textbf{[Question]}

$\langle\!\langle$ question $\rangle\!\rangle$

\textbf{[Response]}

$\langle\!\langle$ Response $\rangle\!\rangle$

\textbf{[Answer]}

$\langle\!\langle correct answer \rangle\!\rangle$

\textbf{[Evaluation]}
\end{tcolorbox}
\subsection{Comparison of PRMs}

Table~\ref{tab:compare_PRM_result} presents a comprehensive comparison of various PRMs across four benchmark datasets: OLY, MATH, GSM8K, and Collegemath.  The models evaluated include Qwen2.5-Math-PRM, Math-Shepherd (ours), Omega, Sample-EDU, and EDU, with parameter sizes ranging from 7B to 72B.  For each dataset, models are grouped according to their parameter sizes to facilitate a fair comparison.  The evaluation is conducted under different sample sizes (2, 4, 8, 16, 32, 64, and 128), allowing for an analysis of performance scaling as the sample size increases.  Bolded values in the table highlight the best-performing model for each sample size within the respective dataset.  This table serves as a supplementary resource for section~\ref{sec:method-section}.

\begin{table*}[t!]
    \centering
    \resizebox{0.75\textwidth}{!}{
    \begin{tabular}{@{}cccccccccc@{}}
        \toprule
        \multirow{2}{*}{Datasets} & \multirow{2}{*}{Models} & \multicolumn{7}{c}{Samples} \\ \cmidrule{3-9}
        &  & 2 & 4 & 8 & 16 & 32 & 64 & 128 \\ \midrule
        & Math-Shepherd-Mistral-7B-PRM & 15.9 & 16.3 & 17.5 & 17.6 & 18.2 & 18.8 & 17.9 \\
        & Qwen2.5-Math-7B-PRM800K & 16 & 18.2 & \underline{19.3} & \underline{19.9} & \underline{20.3} & \underline{21.3} & \underline{22.7} \\
        & Qwen2.5-Math-PRM-7B & \textbf{17.9} & \textbf{20.7} & \textbf{23} & \textbf{23.6} & \textbf{24.6} & \textbf{25.8} & \textbf{28.9} \\
        & Math-Shepherd-7B & 16.9 & 16.4 & 15.1 & 15.1 & 15.4 & 13.9 & 13.8 \\
        & Omega-7B  & 14.5 & 15.3 & 16 & 17.5 & 17.5 & 16.9 & 17.9 \\
        & Sample-EDU-7B  & \underline{17.5} & 18.1 & 18.7 & 18.2 & 19.1 & 19.1 & 20.1 \\
        & EDU-7B  & 16 & \underline{19.4} & 18.4 & 18.2 & 19.7 & 19.4 & 20 \\
        \cmidrule{2-9}
        & Qwen2.5-Math-RM-72B & \textbf{19.4} & 21.8 & 24.4 & 25.5 & 27.4 & \underline{29.2} & \underline{30.4} \\
        & Qwen2.5-Math-PRM-72B & 18.8 & \underline{21.9} & \underline{24.7} & \underline{25.8} & \underline{27} & 28.6 & 29.3 \\
        & Math-Shepherd-72B  & 18.8 & 20.4 & 21.9 & 22.4 & 23.6 & 24.7 & 26.7 \\
        & Omega-72B  & 18.7 & 20.7 & 21.1 & 22.5 & 24.6 & 24.4 & 25.5 \\
        & Sample-EDU-72B  & 18.8 & 21 & 22.2 & 22.4 & 23.6 & 24.1 & 27 \\
        \multirow{-12}{*}{\textbf{OLY}} & EDU-72B  & \textbf{19.4} & \textbf{22.4} & \textbf{25.5} & \textbf{26.7} & \textbf{27.6} & \textbf{30.2} & \textbf{32.7} \\
        \midrule
        & Math-Shepherd-Mistral-7B-PRM & 43.7 & 45.0 & 45.6 & 46.3 & 46.5 & 46.2 & 46.5 \\
        & Qwen2.5-Math-7B-PRM800K & \underline{45.8} & \underline{48.2} & \underline{50.1} & \underline{50.7} & \underline{51} & \underline{51.2} & 51 \\  
        & Qwen2.5-Math-PRM-7B & \textbf{47.4} & \textbf{51.3} & \textbf{54.8} & \textbf{58.2} & \textbf{60.9} & \textbf{62.5} & \textbf{64.6} \\ 
        & Math-Shepherd-7B & 43.8 & 44.8 & 45.2 & 45.5 & 46.2 & 46.2 & 46.1 \\
        & Omega-7B  & 43.4 & 43.7 & 44.5 & 45.6 & 46.8 & 47.6 & 48.5 \\
        & Sample-EDU-7B  & 44 & 46.5 & 47.6 & 48.4 & 49.7 & 50.1 & 50.4 \\
        & EDU-7B  & 44 & 46.3 & 47.7 & 48.9 & 49.6 & 50.6 & \underline{51.3} \\
        \cmidrule{2-9}
        & Qwen2.5-Math-RM-72B & \underline{48.6} & \textbf{54} & \textbf{57.8} & \textbf{62.0} & \textbf{65.4} & \textbf{67.9} & \textbf{70.0} \\
        & Qwen2.5-Math-PRM-72B & 47.2 & 51.5 & 54.8 & 57.9 & 60.5 & 61.7 & 63.6 \\
        & Math-Shepherd-72B  & 47 & 50.9 & 54.4 & 57.1 & 59 & 60.4 & 61.7 \\
        & Omega-72B  & 48 & 52.1 & 54.7 & 57.4 & 59.7 & 61.4 & 62.4 \\
        & Sample-EDU-72B  & 46.9 & 50.4 & 53.8 & 56.5 & 58.8 & 60.3 & 61.8 \\
        \multirow{-12}{*}{\textbf{MATH}} & EDU-72B  & \textbf{48.9} & \underline{53.9} & \underline{57.2} & \underline{61.3} & \underline{62.9} & \underline{64.7} & \underline{65.5} \\
        \midrule
        & Math-Shepherd-Mistral-7B-PRM & \underline{84.7} & 85.2 & 85.4 & 86 & 84.7 & 84.8 & 84.8 \\
        & Qwen2.5-Math-7B-PRM800K & 84.3 & \underline{86.1} & \underline{87} & 87.2 & \underline{87.6} & \underline{88.1} & \underline{87.8} \\
        & Qwen2.5-Math-PRM-7B & \textbf{85.6} & \textbf{87} & \textbf{88.6} & \textbf{88.6} & \textbf{88.9} & \textbf{89.3} & \textbf{89.3} \\
        & Math-Shepherd-7B & 83.3 & 83 & 83.2 & 83.4 & 83 & 83.1 & 82.6 \\
        & Omega-7B  & 82.9 & 83.2 & 83.4 & 83.7 & 85 & 85 & 85.7 \\
        & Sample-EDU-7B  & 82.6 & 82.5 & 82.3 & 82.6 & 83 & 83.4 & 83.5 \\
        & EDU-7B  & 83.9 & 84 & 83.7 & 84.8 & 85.4 & 86.5 & 86.7 \\
        \cmidrule{2-9}
        & Qwen2.5-Math-RM-72B & \textbf{87.3} & \underline{89.7} & \textbf{91.1} & \textbf{91.9} & \textbf{92.3} & \textbf{92.6} & \textbf{92.7} \\
        & Qwen2.5-Math-PRM-72B & 86.4 & 87.7 & 88.7 & 88.9 & 89.3 & 89.9 & 90.3 \\
        & Math-Shepherd-72B  & 86.1 & 87.6 & 88.3 & 88.1 & 88 & 88.6 & 89.5 \\
        & Omega-72B  & 85.4 & 86.3 & 87.6 & 88.6 & 89.2 & 90 & 90.1 \\
        & Sample-EDU-72B  & 85.5 & 87.1 & 87.6 & 87.6 & 87.9 & 88.2 & 88.1 \\
        \multirow{-12}{*}{\textbf{GSM8K}} & EDU-72B  & \underline{87} & \textbf{89.8} & \underline{90.6} & \underline{91.8} & \underline{92.1} & \underline{92} & \underline{91.5} \\
        \midrule
        & Math-Shepherd-Mistral-7B-PRM & \underline{11.8} & 11.8 & 11.8 & 11.6 & 11.7 & 11.8 & 11.8 \\
        & Qwen2.5-Math-7B-PRM800K & 11.7 & \underline{11.9} & 11.8 & 11.6 & 11.6 & 11.5 & 11.6 \\
        & Qwen2.5-Math-PRM-7B & \textbf{11.9} & \textbf{12.3} & \textbf{12.7} & \textbf{13.0} & \textbf{13.2} & \textbf{13.6} & \textbf{14.1} \\
        & Math-Shepherd-7B & 11.5 & 11.8 & 11.9 & 11.9 & 11.8 & 11.9 & 11.9 \\
        & Omega-7B  & 11.7 & 11.6 & 11.7 & 11.8 & 12 & 11.9 & 12.1 \\
        & Sample-EDU-7B  & 11.6 & 12 & \underline{12} & \underline{12.3} & \underline{12.3} & \underline{12.5} & \underline{12.6} \\
        & EDU-7B  & 11.6 & 11.7 & 11.6 & 11.6 & 12.1 & 12 & 12.2 \\
        \cmidrule{2-9}
        & Qwen2.5-Math-RM-72B & \underline{12.1} & \underline{12.6} & \underline{13.3} & \underline{13.9} & \textbf{14.5} & \textbf{15.1} & \textbf{15.7} \\
        & Qwen2.5-Math-PRM-72B & 12 & 12.3 & 12.6 & 12.9 & 13.1 & 13 & 13.2 \\
        & Math-Shepherd-72B  & 12 & 12.5 & 13.2 & 13.8 & 13.8 & 14.3 & 14.8 \\
        & Omega-72B  & 12 & 12.4 & 13.2 & 13.5 & 13.9 & 14.3 & 14.8 \\
        & Sample-EDU-72B  & 11.8 & 12.5 & 12.9 & 13.4 & 13.7 & 14.1 & 14.5 \\
        \multirow{-12}{*}{\textbf{Collegemath}} & EDU-72B  & \textbf{12.3} & \textbf{12.9} & \textbf{13.4} & \textbf{14.1} & \underline{14.4} & \underline{14.9} & \underline{15.5} \\
        \bottomrule
    \end{tabular}
    }
    \caption{Comparison of performance across different datasets (OLY, MATH, GSM8K, and Collegemath) and various PRMs (including Qwen2.5-Math-PRM, Math-Shepherd (ours), Omega, Sample-EDU, and EDU with 7B and 72B parameters, Qwen2.5-Math-7B-PRM800K, Qwen2.5-Math-72B-PRM, Math-Shepherd-Mistral-7B-PRM) under different sample sizes (2, 4, 8, 16, 32, 64, and 128). Models are grouped by parameter size within each dataset. The \textbf{bold} values indicate the highest performance score in each column for the corresponding dataset, and the \underline{underlined} values denote the second highest score.}
    \label{tab:compare_PRM_result}
\end{table*}

\subsection{Performance Comparison of EDU-Based Sample Methods}

Table~\ref{tab:math8_performance} and Table~\ref{tab:oly8_performance} summarize the performance of EDU sampling, P-EDU, and MCTS-EDU methods on the MATH and OLY datasets, respectively, under varying entropy thresholds with a fixed maximum branch number of 8. Each table reports both the accuracy (\%) and the average number of tokens consumed for each method and entropy setting.

The results illustrate several key trends:

\begin{itemize}
    \item For both datasets, increasing the entropy threshold generally leads to a reduction in average token usage, but this is often accompanied by a decrease in accuracy.
    \item The P-EDU Sampling, which incorporates entropy-based pruning, can sometimes outperform the standard EDU Sampling depending on the underlying PRM’s ability to identify and prune low-confidence branches.
    \item The accuracy improvement of MCTS-EDU is constrained by the rollout depth; with limited rollout steps, its accuracy does not continue to increase with higher token counts.
\end{itemize}

These tables provide a comprehensive overview of how entropy-based branching and pruning strategies affect the balance between accuracy and token efficiency across different reasoning methods.

\begin{table*}[!t]
  \centering
  \resizebox{0.95\textwidth}{!}{
  \begin{tabular}{lccccccccc}
    \toprule
    \multicolumn{1}{c}{\multirow{2}{*}{Method}} & \multicolumn{9}{c}{Entropy} \\
    \cmidrule(lr){2-10}
    & 0.8 & 1.0 & 1.2 & 1.4 & 1.6 & 1.8 & 2.0 & 2.2 & 2.4 \\
    \midrule
    EDU-7B & 47.7 & 47.8 & 47.5 & 47.2 & 46.1 & 46.0 & 45.7 & 42.8 & 42.0 \\
    EDU-72B & 58.1 & 57.8 & 57.2 & 57.1 & 56.2 & 54.4 & 51.1 & 51.1 & 49.4 \\
    \midrule
    P-EDU-0.2 & 57.4 & 57.1 & 56.7 & 56.3 & 55.9 & 54.4 & 53.6 & 50.3 & 48.2 \\
    P-EDU-0.3 & 55.6 & 55.5 & 55.5 & 55.1 & 55.2 & 53.8 & 53.2 & 49.8 & 48.6 \\
    P-EDU-0.4 & 52.2 & 52.7 & 53.5 & 52.4 & 53.1 & 52.0 & 52.5 & 48.9 & 48.0 \\
    \midrule
    MCTS-EDU (1-step) & 48.7 & 48.8 & 48.3 & 48.7 & 47.9 & 46.7 & 48.7 & 45.6 & 45.5 \\
    MCTS-EDU (2-step) & 53.2 & 53.2 & 53.6 & 52.9 & 52.5 & 52.2 & 51.8 & 48.7 & 47.8 \\
    MCTS-EDU (3-step) & 57.2 & 56.6 & 56.6 & 55.9 & 55.6 & 54.3 & 53.6 & 50.7 & 49.2 \\
    \midrule
    \textit{EDU Average Token} & 3047 & 3012 & 2988 & 2927 & 2818 & 2650 & 2082 & 2147 & 1880 \\
    \midrule
    \textit{P-EDU-0.2 Average Token} & 3024 & 2988 & 2966 & 2898 & 2769 & 2598 & 2026 & 2074 & 1815 \\
    \textit{P-EDU-0.3 Average Token} & 2434 & 2533 & 2611 & 2610 & 2537 & 2393 & 1904 & 1935 & 1705 \\
    \textit{P-EDU-0.4 Average Token} & 1711 & 1780 & 1875 & 1888 & 1896 & 1835 & 1594 & 1577 & 1405 \\
    \midrule
    \textit{MCTS-EDU (1-step) Average Token} & 1026 & 1010 & 1009 & 997 & 998 & 975 & 937 & 920 & 869 \\
    \textit{MCTS-EDU (2-step) Average Token} & 1863 & 1849 & 1834 & 1818 & 1782 & 1710 & 1464 & 1482 & 1347 \\
    \textit{MCTS-EDU (3-step) Average Token} & 3046 & 3012 & 2979 & 2915 & 2788 & 2616 & 2030 & 2098 & 1880 \\
    \bottomrule
  
  \end{tabular}
  }
  \caption{Accuracy and Average Token Usage of EDU Sampling, P-EDU, and MCTS-EDU Methods on the MATH Dataset Across Different Entropy Thresholds (Max Branches = 8). Higher entropy values correspond to later branching and fewer tokens. The table reports both accuracy (\%) and average token count for each method and threshold.}
  \label{tab:math8_performance}
\end{table*}

\begin{table*}[!t]
  \centering
  \resizebox{0.95\textwidth}{!}{
  \begin{tabular}{lccccccccc}
    \toprule
    \multicolumn{1}{c}{\multirow{2}{*}{Method}} & \multicolumn{9}{c}{Entropy} \\
    \cmidrule(lr){2-10}
    & 0.8 & 1.0 & 1.2 & 1.4 & 1.6 & 1.8 & 2.0 & 2.2 & 2.4 \\
    \midrule
    EDU-7B & 21.5 & 20.8 & 20.0 & 18.8 & 18.3 & 20.0 & 21.3 & 20.0 & 19.4 \\
    EDU-72B & 26.9 & 26.5 & 25.5 & 26.9 & 25.1 & 25.4 & 26.7 & 26.2 & 25.7 \\
    \midrule
    P-EDU-0.2 & 27.0 & 27.6 & 25.2 & 24.8 & 25.4 & 25.2 & 25.9 & 25.4 & 26.5 \\
    P-EDU-0.3 & 25.5 & 26.4 & 24.4 & 24.2 & 24.2 & 24.6 & 25.6 & 24.7 & 25.8 \\
    P-EDU-0.4 & 23.3 & 24.1 & 22.5 & 22.1 & 23.1 & 22.2 & 25.1 & 24.4 & 24.4 \\
    \midrule
    MCTS-EDU (1-step) & 21.8 & 22.8 & 20.6 & 21.6 & 21.0 & 20.2 & 21.7 & 20.2 & 21.7 \\
    MCTS-EDU (2-step) & 24.8 & 24.6 & 23.8 & 24.2 & 23.7 & 22.9 & 23.8 & 24.7 & 23.5 \\
    MCTS-EDU (3-step) & 26.0 & 26.1 & 24.3 & 24.5 & 24.3 & 24.6 & 25.1 & 24.9 & 25.0 \\
    \midrule
    \textit{EDU Average Token} & 3973 & 3961 & 3980 & 4030 & 4010 & 4013 & 3924 & 3801 & 3576 \\
    \midrule
    \textit{P-EDU-0.2 Average Token} & 3948 & 3930 & 3937 & 3979 & 3946 & 3926 & 3853 & 3702 & 3492 \\
    \textit{P-EDU-0.3 Average Token} & 3122 & 3227 & 3352 & 3417 & 3474 & 3488 & 3499 & 3399 & 3236 \\
    \textit{P-EDU-0.4 Average Token} & 2260 & 2721 & 2844 & 2916 & 2962 & 3016 & 3082 & 3095 & 2936 \\
    \midrule
    \textit{MCTS-EDU (1-step) Average Token} & 1449 & 1430 & 1437 & 1437 & 1451 & 1428 & 1432 & 1388 & 1347 \\
    \textit{MCTS-EDU (2-step) Average Token} & 2567 & 2543 & 2561 & 2573 & 2576 & 2574 & 2541 & 2532 & 2389 \\
    \textit{MCTS-EDU (3-step) Average Token} & 2972 & 3961 & 3981 & 4025 & 4014 & 4009 & 3909 & 3792 & 3547 \\
    \bottomrule
  \end{tabular}
  }
  \caption{Accuracy (\%) Comparison of EDU Sampling, P-EDU Sampling, and MCTS-EDU on OLY Dataset under Different Entropy Values (Max Branches = 8)}
  \label{tab:oly8_performance}
\end{table*}

\subsection{Comprehensive Comparison of EDU Sampling on MATH and OLY Datasets by different Maximum branch}

Table~\ref{tab:math_oly_combined} presents a detailed comparison of several branching strategies—HT Sampling, EDU Sampling, P-EDU Sampling, and MCTS Sampling—on both the MATH and OLY datasets as the maximum allowed number of branches varies from 1 to 64. The table includes three main metrics: accuracy (\%) using the 72B model, total tokens consumed (in millions), and average tokens per problem for each method and branch setting.

Key observations include:
\begin{itemize}
    \item Increasing the maximum branch number generally leads to higher accuracy for most methods, but also significantly increases token usage.
    \item EDU Sampling and P-EDU Sampling demonstrate better token efficiency compared to HT Sampling, especially at higher branch limits.
    \item MCTS Sampling’s accuracy plateaus or even drops at higher branch numbers, but its token usage remains relatively low due to its targeted search mechanism.
    \item OLY dataset results show lower overall accuracy compared to MATH, but similar scaling trends in token usage and efficiency.
\end{itemize}

This table provides a comprehensive overview of how different branching and sampling strategies scale with computational resources, highlighting the trade-offs between accuracy gains and token consumption.

\begin{table*}[!t]
  \centering
  \resizebox{0.95\textwidth}{!}{
  \renewcommand{\arraystretch}{1.0}
  \small  
  \begin{tabular}{l@{\hspace{2pt}}*{7}{r@{\hspace{2pt}}}@{\hspace{8pt}}*{7}{r@{\hspace{2pt}}}}
    \toprule
    & \multicolumn{7}{c@{\hspace{8pt}}}{\textbf{MATH Dataset}} & \multicolumn{7}{c}{\textbf{OLY Dataset}} \\
    \cmidrule(lr){2-8} \cmidrule(l){9-15}
    \textbf{Method} & \textbf{1} & \textbf{2} & \textbf{4} & \textbf{8} & \textbf{16} & \textbf{32} & \textbf{64} & \textbf{1} & \textbf{2} & \textbf{4} & \textbf{8} & \textbf{16} & \textbf{32} & \textbf{64} \\
    \midrule
    \multicolumn{15}{l}{\textbf{Performance (\%) - 72B Model}} \\
    \addlinespace[2pt]
    HT Sampling & 42.2 & 48.9 & 53.9 & 57.2 & 61.3 & 62.9 & 64.7 & 14.2 & 19.4 & 22.4 & 25.5 & 26.7 & 27.6 & 30.2 \\
    EDU Sampling & 41.8 & 50.7 & 55.0 & 57.4 & 62.4 & 64.7 & 67.3 & 20.2 & 21.7 & 24.8 & 26.7 & 28.9 & 31.7 & \textbf{33.2} \\
    \addlinespace[1pt]
    P-EDU (0.2) & 41.8 & 46.3 & 51.1 & 57.1 & 60.8 & 63.2 & 65.2 & 20.2 & 21.5 & 25.1 & 25.9 & 28.8 & 32.1 & 32.2 \\
    P-EDU (0.3) & 41.8 & 46.3 & 51.1 & 55.5 & 59.7 & 61.8 & 63.7 & 20.2 & 21.5 & 24.7 & 25.6 & 28.1 & 30.9 & 30.0 \\
    P-EDU (0.4) & 41.8 & 46.3 & 50.8 & 52.7 & 56.0 & 57.4 & 59.2 & 20.2 & 21.5 & 23.1 & 25.1 & 24.4 & 26.2 & 27.8 \\
    \addlinespace[1pt]
    MCTS (1) & 41.8 & 46.3 & 50.4 & 48.8 & 48.6 & 47.6 & 47.8 & 20.2 & 21.5 & 22.7 & 21.7 & 20.5 & 21.2 & 22.1 \\
    MCTS (2) & 41.8 & 46.3 & 51.1 & 53.2 & 53.7 & 54.2 & 53.4 & 20.2 & 21.5 & 25.3 & 23.8 & 23.1 & 23.0 & 25.5 \\
    MCTS (3) & 41.8 & 46.3 & 51.2 & 56.6 & 57.2 & 55.9 & 56.8 & 20.2 & 21.5 & 25.3 & 25.1 & 25.0 & 24.8 & 26.4 \\
    \midrule
    \multicolumn{15}{l}{\textbf{Token Usage Statistics}} \\
    \addlinespace[2pt]
    \multicolumn{15}{l}{\textit{Total Tokens (M)}} \\
    HT Sampling & 2.65 & 5.28 & 10.7 & 21.7 & 43.3 & 86.5 & 173 & 0.58 & 1.12 & 2.23 & 4.45 & 8.92 & 17.9 & 35.7 \\
    EDU Sampling & 0.49 & 0.93 & 1.80 & 3.66 & 7.38 & 14.8 & 29.9 & 0.49 & 0.93 & 1.80 & 3.66 & 7.38 & 14.8 & 29.9 \\
    \addlinespace[2pt]
    \multicolumn{15}{l}{\textit{Average Tokens per Problem}} \\
    BON Sampling & 530 & 1,056 & 2,146 & 4,338 & 8,650 & 17,306 & 34,623 & 853 & 1,655 & 3,298 & 6,591 & 13,213 & 26,489 & 52,848 \\
    EDU Sampling & 511 & 700 & 946 & 2,988 & 5,980 & 11,882 & 23,546 & 643 & 1,107 & 2,034 & 3,749 & 7,153 & 15,050 & 30,524 \\
    \addlinespace[1pt]
    P-EDU (0.2) & 511 & 700 & 937 & 2,031 & 3,777 & 7,753 & 22,867 & 643 & 1,107 & 2,034 & 3,930 & 7,570 & 15,050 & 30,524 \\
    P-EDU (0.3) & 511 & 700 & 919 & 1,908 & 3,415 & 6,824 & 15,174 & 643 & 1,107 & 1,938 & 3,227 & 6,365 & 11,710 & 18,565 \\
    P-EDU (0.4) & 511 & 700 & 874 & 1,597 & 2,569 & 4,591 & 6,896 & 643 & 1,107 & 1,660 & 2,323 & 3,804 & 5,827 & 8,540 \\
    \addlinespace[1pt]
    MCTS (1) & 511 & 700 & 787 & 936 & 933 & 955 & 1,053 & 643 & 1,107 & 1,339 & 1,432 & 1,475 & 1,480 & 1,489 \\
    MCTS (2) & 511 & 700 & 639 & 1,465 & 1,666 & 1,681 & 2,038 & 643 & 1,107 & 2,046 & 2,541 & 2,762 & 2,825 & 2,931 \\
    MCTS (3) & 511 & 700 & 946 & 2,037 & 2,633 & 2,959 & 3,963 & 643 & 1,107 & 2,048 & 3,909 & 4,932 & 5,423 & 5,683 \\
    \bottomrule
  \end{tabular}
  }
  \caption{Accuracy and Token Usage Statistics for HT Sampling, EDU Sampling, P-EDU Sampling, and MCTS Sampling across Different Maximum Branch Numbers (1–64) on the MATH and OLY Datasets. The table reports accuracy (\%) for the 72B model, total tokens consumed (in millions), and average tokens used per problem for each configuration, illustrating the trade-offs between performance and computational cost as the branch limit increases.}
  \label{tab:math_oly_combined}
\end{table*}

\subsection{Multi-Level Pruning Impact on PRM Score Distribution} 

\begin{figure*}[!t]
    \centering
    \includegraphics[width=0.8\textwidth]{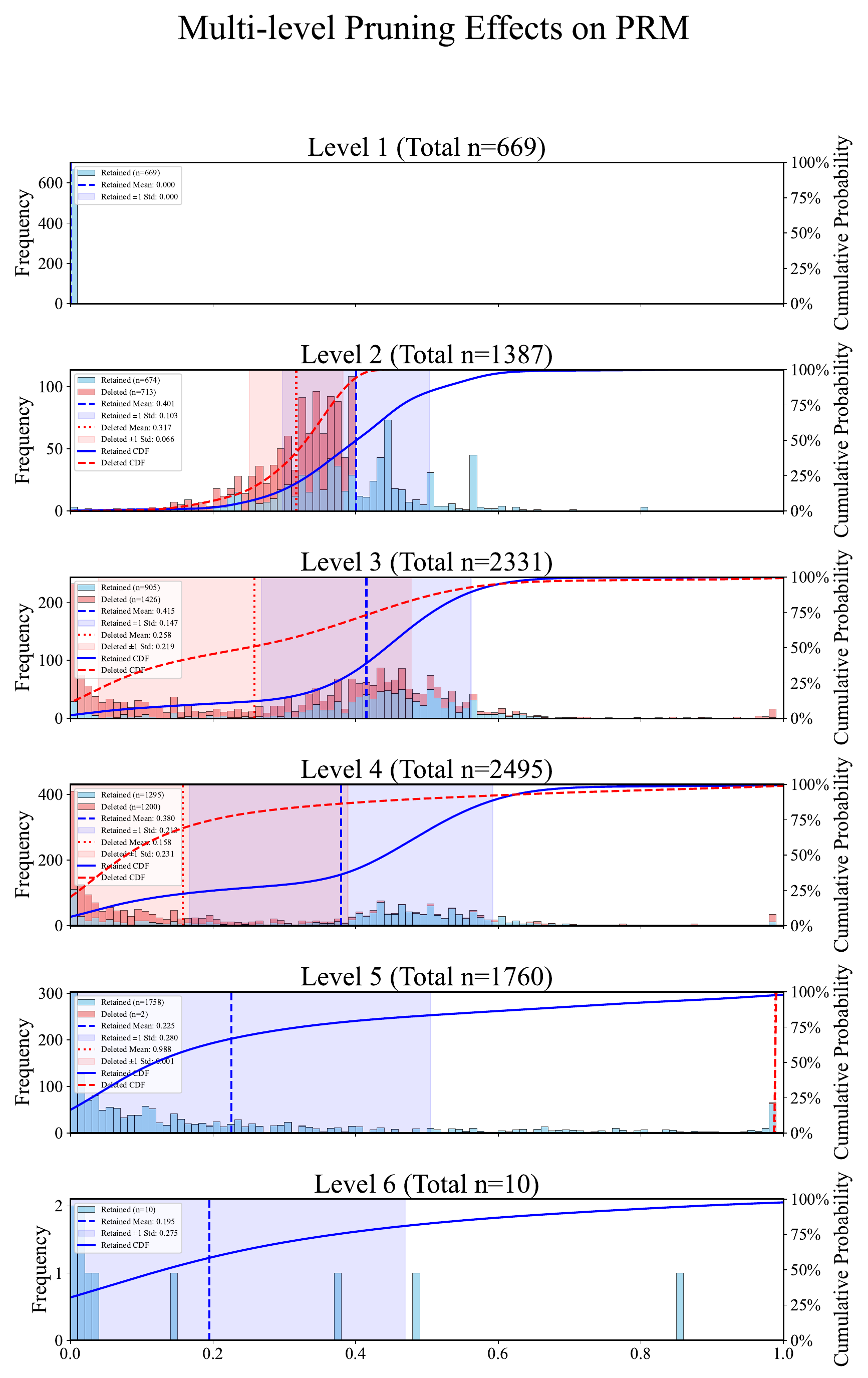}
    \caption{Multi-level Pruning Effects on PRM. This visualization presents the distribution of PRM scores across six levels (1 to 6) for a large model, illustrating the effect of threshold-based pruning on node retention and deletion. Each panel includes a cumulative distribution function (CDF) comparing retained and deleted nodes, along with frequency histograms showing their counts, and displays the mean PRM scores for both groups.}
    \label{fig:zj}
\end{figure*}

This figure~\ref{fig:zj} illustrates the effects of multi-level threshold-based pruning on PRM scores for a large model. The visualization covers six pruning levels (from 1 to 6), showing how the distribution of PRM scores changes as nodes are either retained or deleted. For each level, the panels display the cumulative distribution functions (CDFs) comparing retained and deleted nodes, as well as frequency histograms indicating their counts. Additionally, the mean PRM scores for both groups are presented, providing insight into the impact of pruning on model performance and node characteristics.

\subsection{Word Frequency Analysis Across Datasets and Branch Configurations}

\begin{figure*}[!t]
    \centering
    \begin{subfigure}[b]{0.45\textwidth}
        \includegraphics[width=\textwidth]{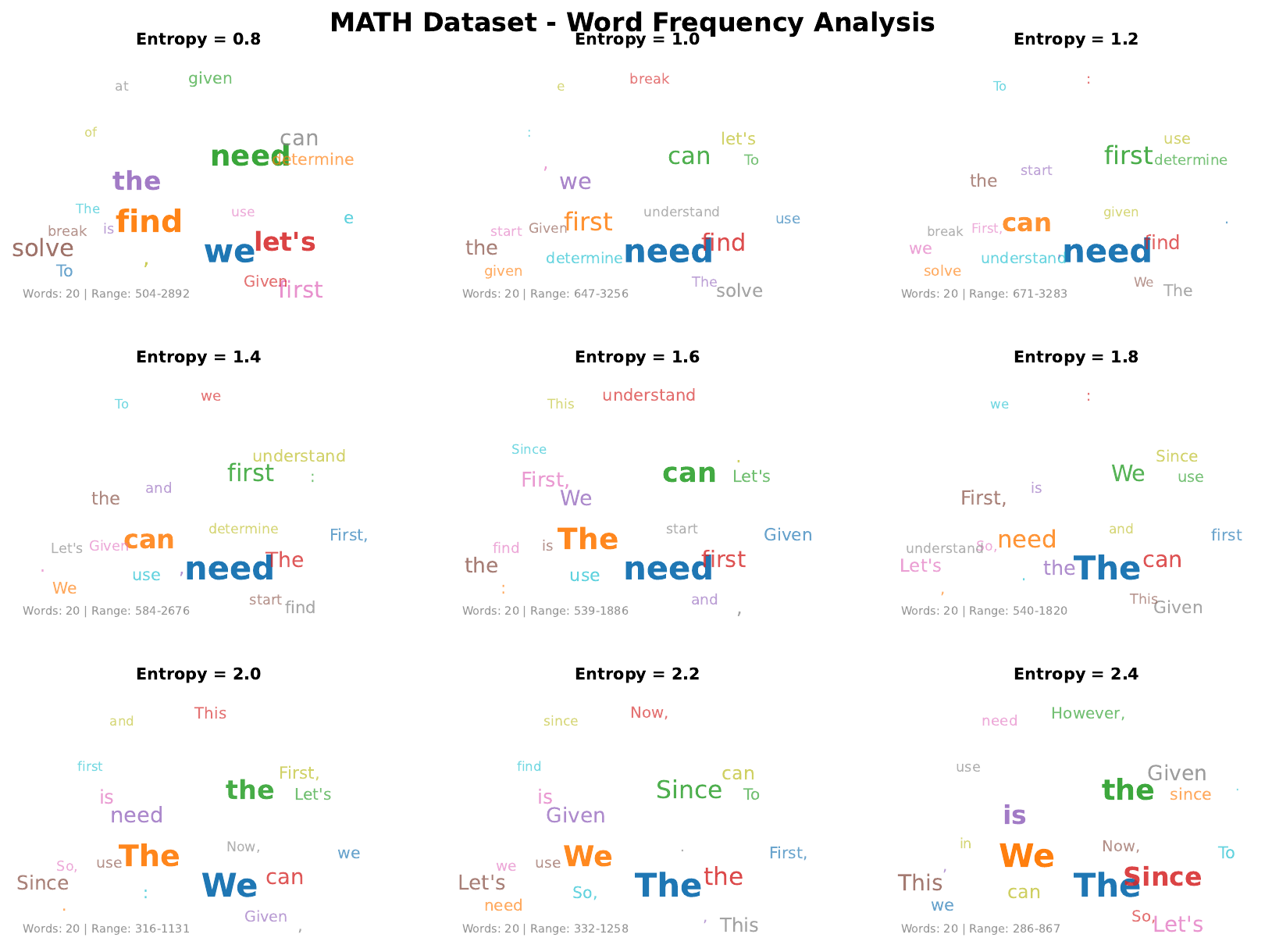}
        \caption{Math@8}
    \end{subfigure}
    \begin{subfigure}[b]{0.45\textwidth}
        \includegraphics[width=\textwidth]{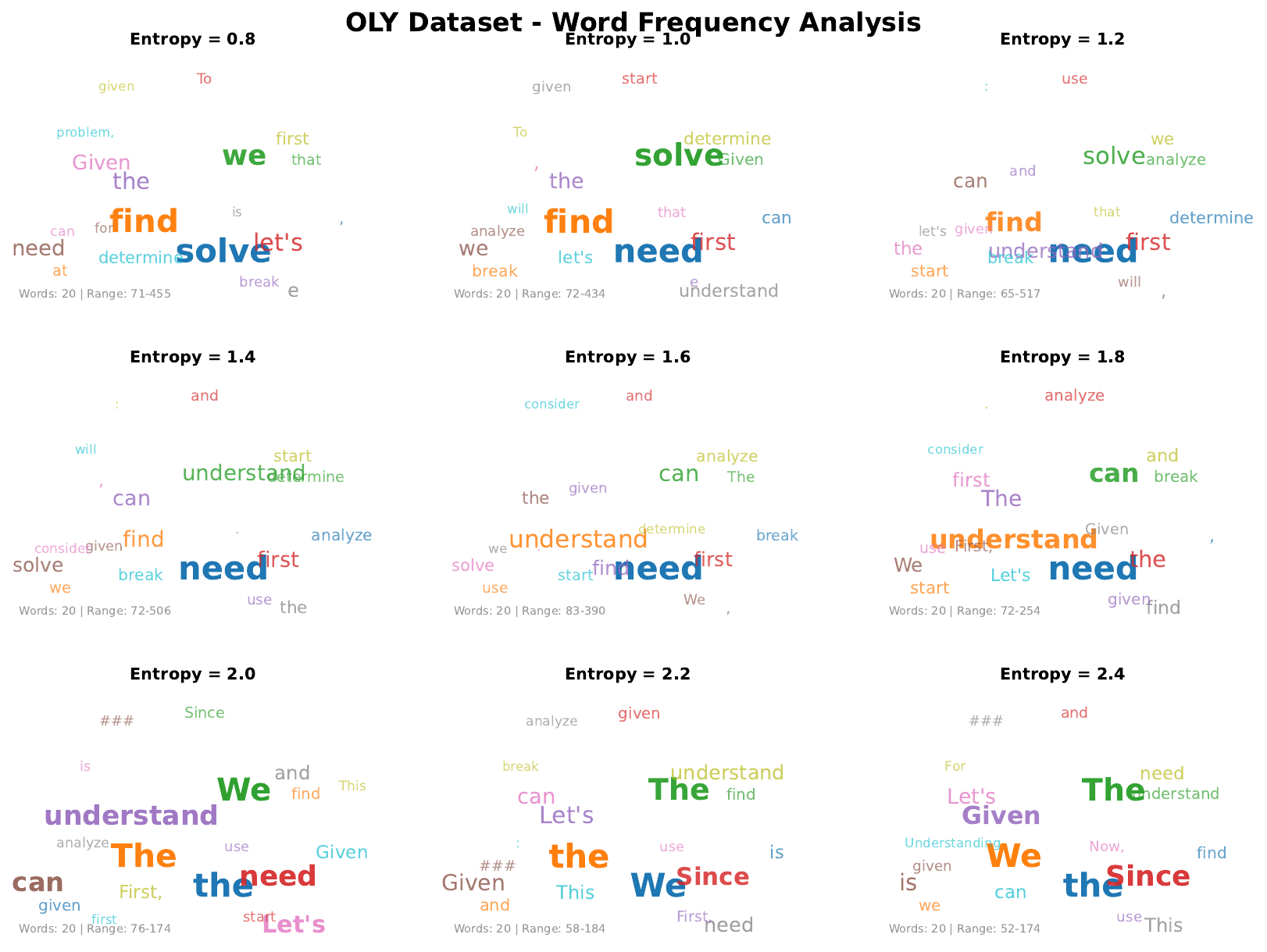}
        \caption{Oly@8}
    \end{subfigure}
    \caption{Word cloud visualizations for the MATH and OLY datasets under different entropy conditions by EDU Sampling, where the maximum branch number is set to 8. The size of each word reflects its frequency in the dataset, with more frequent words shown in larger font.}
    \label{fig:word_frequency_dataset}
\end{figure*}

Figure~\ref{fig:word_frequency_dataset} presents word cloud visualizations for the MATH and OLY datasets under different entropy conditions, with the maximum branch number set to 8. In these visualizations, the size of each word corresponds to its frequency within the dataset, allowing for an intuitive comparison of commonly used terms across different entropy settings.

\begin{figure*}[!t]
    \centering
    \begin{subfigure}[b]{0.45\textwidth}
        \includegraphics[width=\textwidth]{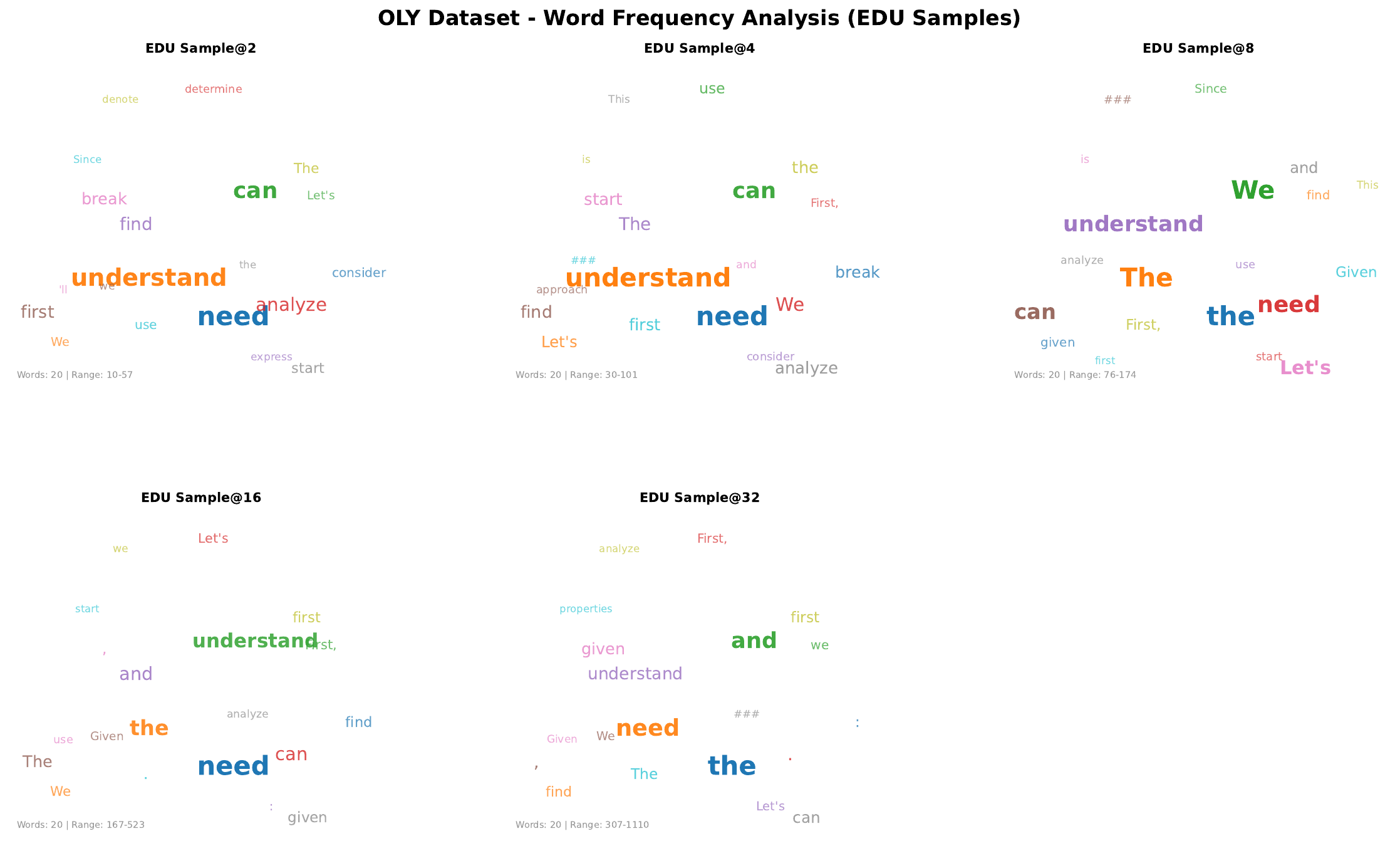}
        \caption{Oly EDU}
    \end{subfigure}
    \begin{subfigure}[b]{0.45\textwidth}
        \includegraphics[width=\textwidth]{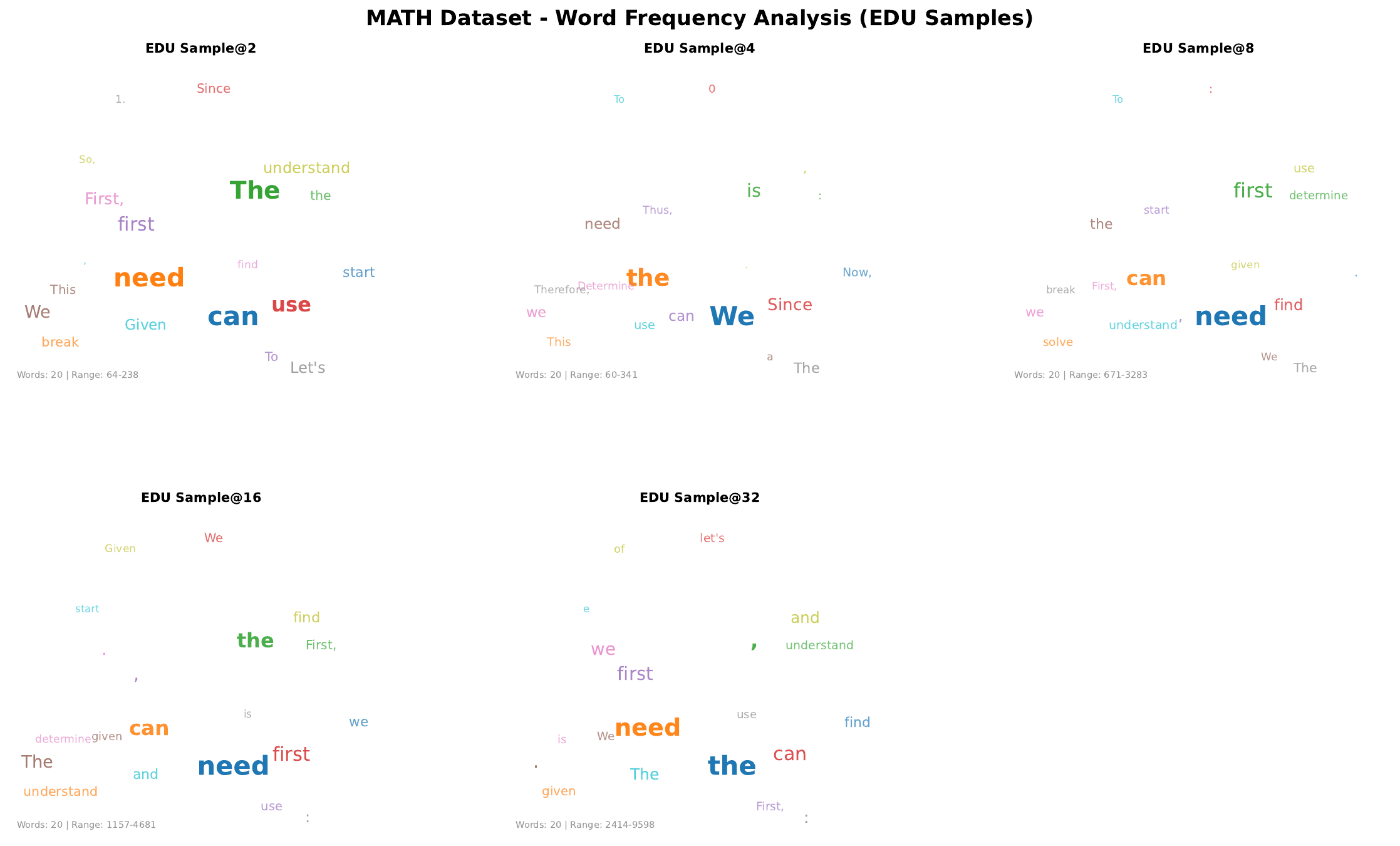}
        \caption{Math EDU}
    \end{subfigure}
    \caption{Word cloud visualizations for Oly and MATH samples under different maximum branch numbers by EDU Sampling. The font size of each word indicates its frequency, with more frequently occurring words shown in larger font.}
    \label{fig:word_frequency_edu}
\end{figure*}

Figure~\ref{fig:word_frequency_edu} shows word cloud visualizations for OLY and MATH samples under varying maximum branch numbers. The font size of each word indicates its frequency, with larger fonts representing words that appear more frequently in the samples. These figures provide insights into the distribution of key terms in educational samples, highlighting differences in word usage patterns across datasets and branching configurations.

\subsection{Illustrative Example of an EDU Sampling}

\begin{figure*}[!t]
    \centering
    \includegraphics[width=0.974\textwidth]{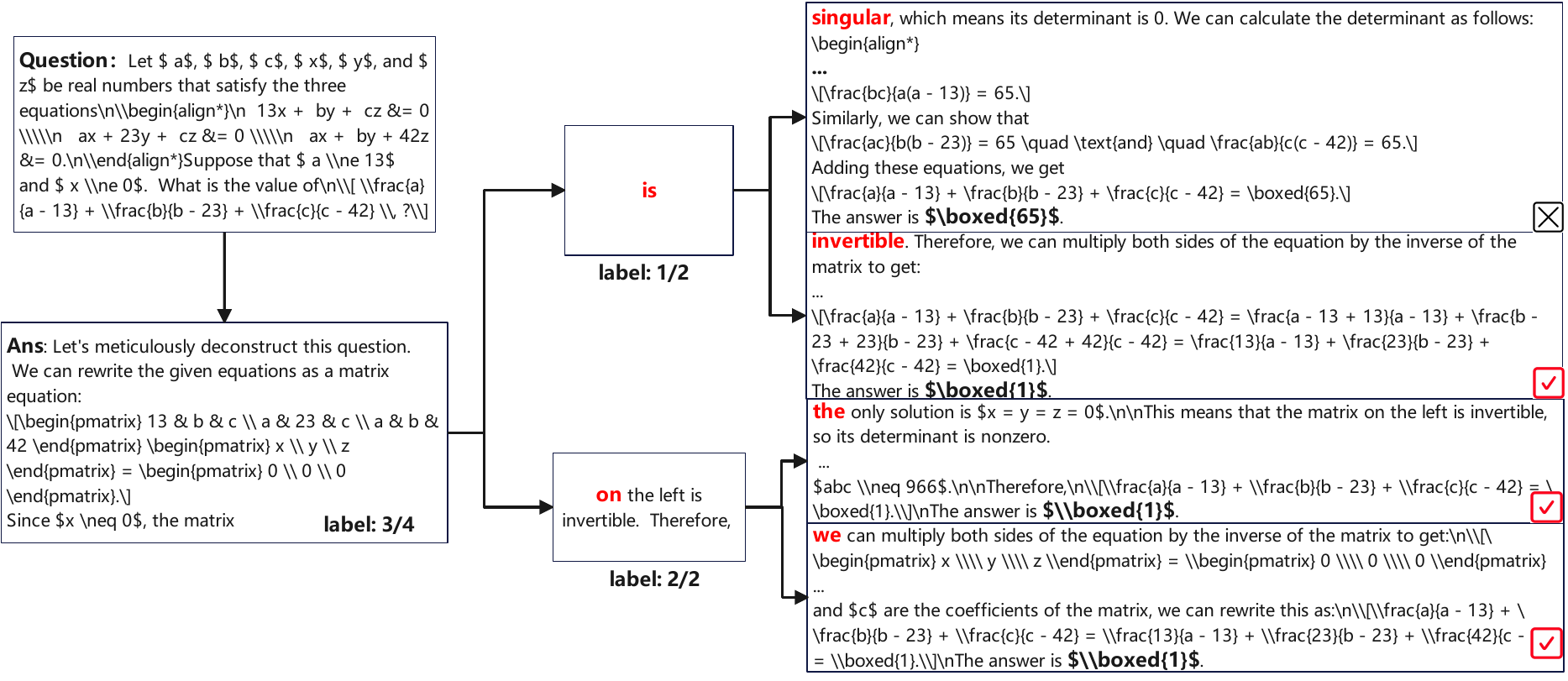}
    \caption{This is a real example of an EDU Sampling, where a selected branch is presented for illustration. The red-colored segments correspond to tokens with entropy values below the predefined threshold. For each step, the Label is derived from the results obtained through backpropagation based on the final outcome.}
    \label{fig:example_of_edu-sample}
\end{figure*}

Figure~\ref{fig:example_of_edu-sample} presents a real example of an EDU Sampling, illustrating the process of branch selection and token evaluation. In this example, a specific branch is highlighted for clarity. The segments shown in red represent tokens whose entropy values fall below the predefined threshold, indicating points of higher confidence during the reasoning process. At each step, the Label is determined through backpropagation from the final solution outcome, providing insight into the contribution of each token to the overall result. This visualization demonstrates how entropy-based selection and backpropagation labeling work together to guide the reasoning trajectory in the EDU Sampling framework.

\subsection{Heatmap Analysis of Node Branch Point Distributions}

\begin{figure*}[!t]
    \centering
    \includegraphics[width=0.95\textwidth]{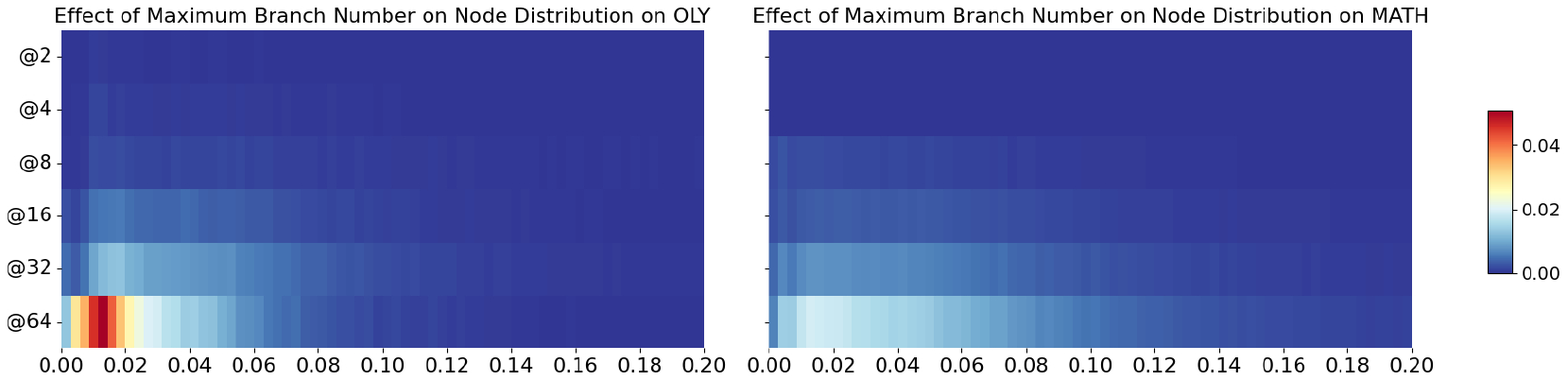}
    \caption{Heatmaps of node distribution under different Maximum Branch Number settings on the OLY and MATH test sets, restricted to the 0–20\% interval of solutions.  Red denotes a higher concentration of nodes in that percentile range, whereas blue denotes a lower concentration.  Relative to MATH, OLY exhibits a more front‑loaded (early‑range) concentration.}
    \label{fig:heatmap_max}
\end{figure*}

\begin{figure*}[!t]
    \centering
    \includegraphics[width=0.95\textwidth]{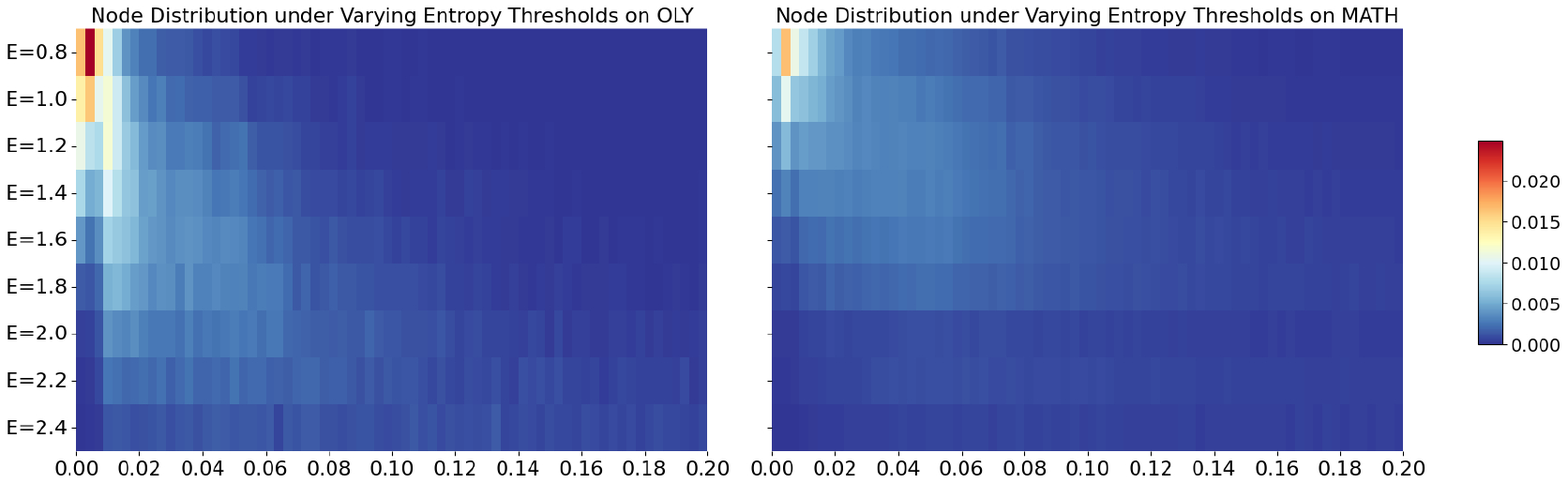}
    \caption{Heatmaps of branch point distribution on the OLY and MATH test sets at a Max Branch Number of 8 under varying entropy thresholds (only the 1–20\% segment shown). Lower entropy thresholds trigger earlier branching, and for any fixed threshold, OLY exhibits earlier branch points than MATH.}
    \label{fig:heatmap_entropy}
\end{figure*}

Figure~\ref{fig:heatmap_max} and Figure~\ref{fig:heatmap_entropy} provide heatmap visualizations of node and branch point distributions under different experimental conditions on the OLY and MATH test sets.

Figure~\ref{fig:heatmap_max} shows the concentration of nodes within the initial 0–20\% interval of solution steps for varying Maximum Branch Number settings. Red regions indicate a higher concentration of nodes, while blue regions represent lower concentrations. Compared to MATH, the OLY test set displays a more front-loaded distribution, with nodes concentrated earlier in the solution process.

Figure~\ref{fig:heatmap_entropy} illustrates branch point distributions at a fixed Maximum Branch Number of 8 under different entropy thresholds, focusing on the 1–20\% segment. Lower entropy thresholds result in earlier branching, and for any given threshold, OLY consistently shows branch points occurring earlier than MATH. These observations highlight structural differences in reasoning trajectories and branching dynamics between the two datasets.


\subsection{Token Count vs. Accuracy Analysis Across Sampling Methods with different entropy}

\begin{figure*}[!t]
    \centering
    \includegraphics[width=0.95\linewidth]{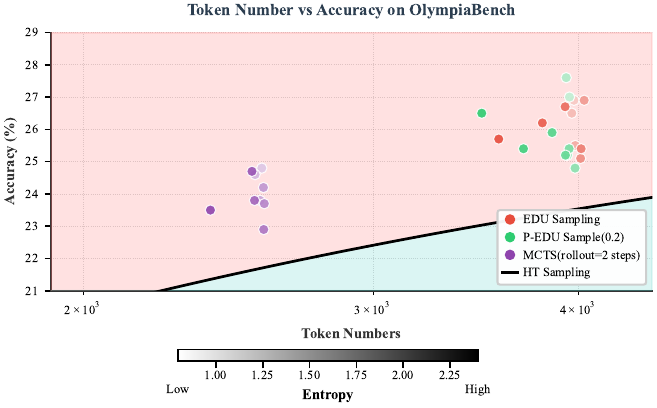}
    \caption{This figure illustrates the relationship between token count and accuracy on the OlympiaBench test set under a Max Branch Number of 8, with the performance of HT Sampling across varying token counts fitted as the baseline. On the MATH test set, most data points for both \textcolor{color9}{EDU Sampling} and \textcolor{color10}{P-EDU(0.2) Sampling} lie above this baseline. Notably, as the entropy threshold increases, token counts decrease alongside a corresponding drop in accuracy.}
    \label{fig:entropy_acc_token_oly}
\end{figure*}

Figure~\ref{fig:entropy_acc_token_oly} illustrates the relationship between token count and accuracy on the OlympiaBench and MATH test sets under a Max Branch Number of 8. The performance of HT Sampling across different token counts is fitted as the baseline for comparison. On the MATH test set, most data points for both EDU Sampling and P-EDU(0.2) Sampling are positioned above this baseline, indicating superior performance in terms of accuracy relative to token count. As the entropy threshold increases, the number of tokens required decreases, but this reduction is accompanied by a corresponding drop in accuracy. Additionally, the MCTS method also exceeds the HT Sampling baseline when the entropy threshold is set lower, further highlighting the impact of entropy-based branching strategies on solution efficiency and accuracy.

\end{document}